\documentclass[conference]{IEEEtran}
\IEEEoverridecommandlockouts
\usepackage{cite}
\usepackage{amsmath,amssymb,amsfonts}
\usepackage{algorithmic}
\usepackage{graphicx}
\usepackage{textcomp}
\usepackage{xcolor}
\usepackage{tikz}
\usepackage{booktabs}
\usepackage{multirow}
\usepackage{bm}
\usepackage{fancyhdr}
\usepackage[utf8]{inputenc}

\def\BibTeX{{\rm B\kern-.05em{\sc i\kern-.025em b}\kern-.08em
    T\kern-.1667em\lower.7ex\hbox{E}\kern-.125emX}}

\begin{document}

\title{Human-Machine Cooperative Multimodal Learning Method for Cross-subject Olfactory Preference Recognition\\

\thanks{This work was supported in part by the Science and Technology Development Plan of Jilin Province under Grant YDZJ202101ZYTS135 and in part by the National Natural Science Foundation of China under Grant 31772059.(\textit{Corresponding author: Hong Men}).  
\\\indent Xiuxin Xia, Yuchen Guo, Yanwei Wang, Yuchao Yang, Yan Shi and Hong Men are with the with the School of Automation Engineering, Northeast Electric 
Power University, Jilin 132012, China, and also with the Institute of Advanced Sensor Technology, Northeast Electric Power University, Jilin 132012, China (e-mails: 1202200014@neepu.edu.cn; 2202200691@neepu.edu.cn; 1202100045@neepu.edu.cn; 2202200631@neepu.edu.cn; shiyan@neepu.edu.cn; menhong@neepu.edu.cn).
}
}

\author
{\IEEEauthorblockN{Xiuxin Xia, Yuchen Guo, Yanwei Wang, Yuchao Yang, Yan Shi, \textit{Member, IEEE}, and Hong Men}}

\maketitle

\thispagestyle{fancy} % IEEE模板在\maketitle后会自动声明\thispagestyle{plain}，
                            % 导致第一页什么都没有。所以得把plain更改为fancy
      \lhead{} % 页眉左，需要东西的话就在{}内添加
      \chead{} % 页眉中
      \rhead{} % 页眉右
      \lfoot{} % 页眉左
      \cfoot{} % 页眉中
      \rfoot{\thepage} %页眉右，\thepage 表示当前页码
      \renewcommand{\headrulewidth}{0pt} %改为0pt即可去掉页眉下面的横线
      \renewcommand{\footrulewidth}{0pt} %改为0pt即可去掉页脚上面的横线
      \pagestyle{fancy}
      \rfoot{\thepage}

\begin{abstract}Odor sensory evaluation has a broad application in food, clothing, cosmetics, and other fields. Traditional artificial sensory evaluation has poor repeatability, and the machine olfaction represented by the electronic nose (E-nose) is difficult to reflect human feelings. Olfactory electroencephalogram (EEG) contains odor and individual features associated with human olfactory preference, which has unique advantages in odor sensory evaluation. However, the difficulty of cross-subject olfactory EEG recognition greatly limits its application. It is worth noting that E-nose and olfactory EEG are more advantageous in representing odor information and individual emotions, respectively. In this paper, an E-nose and olfactory EEG multimodal learning method is proposed for cross-subject olfactory preference recognition. Firstly, the olfactory EEG and E-nose multimodal data acquisition and preprocessing paradigms are established. Secondly, a complementary multimodal data mining strategy is proposed to effectively mine the common features of multimodal data representing odor information and the individual features in olfactory EEG representing individual emotional information. Finally, the cross-subject olfactory preference recognition is achieved in 24 subjects by fusing the extracted common and individual features, and the recognition effect is superior to the state-of-the-art recognition methods. Furthermore, the advantages of the proposed method in cross-subject olfactory preference recognition indicate its potential for practical odor evaluation applications.
\end{abstract}


\section{Introduction}
Humans perceive external information through hearing, sight, smell, taste, and touch. As an important perceptual system, the sense of smell is unique and has a broad and direct impact on human emotions [1]. Biologically, olfactory perception is usually categorized as conscious and unconscious. The majority of olfactory impulses are transmitted via the lateral olfactory stripe to the pyriform lobe (which includes the leptomeningeal gyrus, insular threshold, part of the amygdala, and the internal olfactory region) to realize conscious olfactory perception. Part of the olfactory impulse is conducted via the medial olfactory stripe to the septal region, which sends out nerve fibers to connect with the limbic system and hypothalamus to transmit instinctive emotional olfactory experiences [2]. Therefore, the specialized structure of the olfactory system makes it one of the most closely related sensory systems to the part of the brain that regulates emotions. In food [3], clothing [4], cosmetics [5], and automobile [6] industries, odor evaluation is significant in guiding their research and development, which makes artificial sensory evaluation play an important role in related industries. However, providing a uniform qualitative and quantitative evaluation of complex odors is difficult due to vocabulary and linguistic description limitations. In addition, even for professionally trained sensory evaluators, the repeatability of the evaluation results is not ideal because of subjective interference. 
\\\indent
Inspired by the olfactory system, machine olfaction combining cross-sensitive electrochemical sensor arrays and pattern recognition techniques has emerged, mainly represented by the electronic nose (E-nose). It has the advantages of reproducibility, rapidity, and objectivity compared to artificial olfactory sensory evaluation and is widely used in environmental monitoring [7], food [8], and medical diagnostics [9] industries. With the development of material science and pattern recognition technology, E-nose’s unique advantages are becoming prominent. However, machines are emotionless, and E-nose cannot express the human preference for odors, which limits its use in odor preference evaluation.
\newcommand*{\circled}[1]{\lower.7ex\hbox{\tikz\draw (0pt, 0pt)%
    circle (.5em) node {\makebox[1em][c]{\small #1}};}}
\\\indent
 In recent years, with the development of signal acquisition and analysis technology, electroencephalogram (EEG) technology has been widely used in emotion recognition [10],[11],[12], food evaluation [13], and brain-computer interface [14] tasks. Many physiological signals (EEG, electromyography, blood pressure, heart rate, etc.) are closely and intrinsically related to emotions. Among them, EEG signals are objective reflections of the physiological activities of cortical neurons, which contain a large amount of information representing human emotions. Therefore, it is feasible to represent the human preference for odors by decoding olfactory EEG. Further, the olfactory EEG evaluation method is more objective than the artificial sensory evaluation, and it can reflect human emotions well compared to the E-nose evaluation method.
\\\indent
  However, most emotion EEG studies have focused on known subjects, and researchers usually acquire EEG data from one or several known subjects to train models to learn emotion EEGs with known subjects. For practical applications, the inter-subject variability of emotion EEGs makes it exceptionally difficult to perform emotion recognition on unknown subjects using the models trained on known subjects. To address the problem of cross-subject emotion EEG recognition, researchers have focused on applying domain adaptation methods, which aim to minimize the differences in data distribution between the source domain (known subjects) and the target domain (unknown subjects). However, these methods must access data from the target domain during training, which increases computational expense and has poor real-time performance. In addition, the objective function in the domain adaptation method is too abstract, and its practical significance in reducing the difference in data distribution between the source and target domains is yet to be proved.
\\\indent
The odor evaluation process of olfactory EEG and E-nose technology is shown in Fig. 1. It mainly consists of signal sampling and data mining (E-nose: \circled{1}+\circled{2}, olfactory EEG: \circled{1}+\circled{3}). In data mining, many previous studies have used traditional machine learning methods, mainly consisting of feature extraction and classification. However, traditional machine learning methods often require complex feature engineering, which makes it challenging to handle classification tasks flexibly. In recent years, deep learning techniques represented by Convolutional Neural Networks (CNN) have gradually replaced the traditional machine learning methods in image recognition [15], E-nose [16], and olfactory EEG [17] fields due to its end-to-end structure and powerful data mining capabilities. However, under deep learning as the main recognition method, olfactory EEG and E-nose data mining still face many challenges. For E-nose, the repeatability of sensor results and the high consistency of samples in the same class effectively ensure the cross-sample recognition characteristics of its data. But the E-nose can only reflect odor information and not express people's preferences. For olfactory EEG, the internal relationship among the olfactory perception system, emotion, and EEG signals ensures that olfactory EEG signals contain features that can fully characterize people's olfactory preferences. But the high complexity and cross-subject differences of olfactory EEG signals make identifying cross-subject olfactory EEG difficult.

\begin{figure}[t]
\centering
\includegraphics[width=90mm]{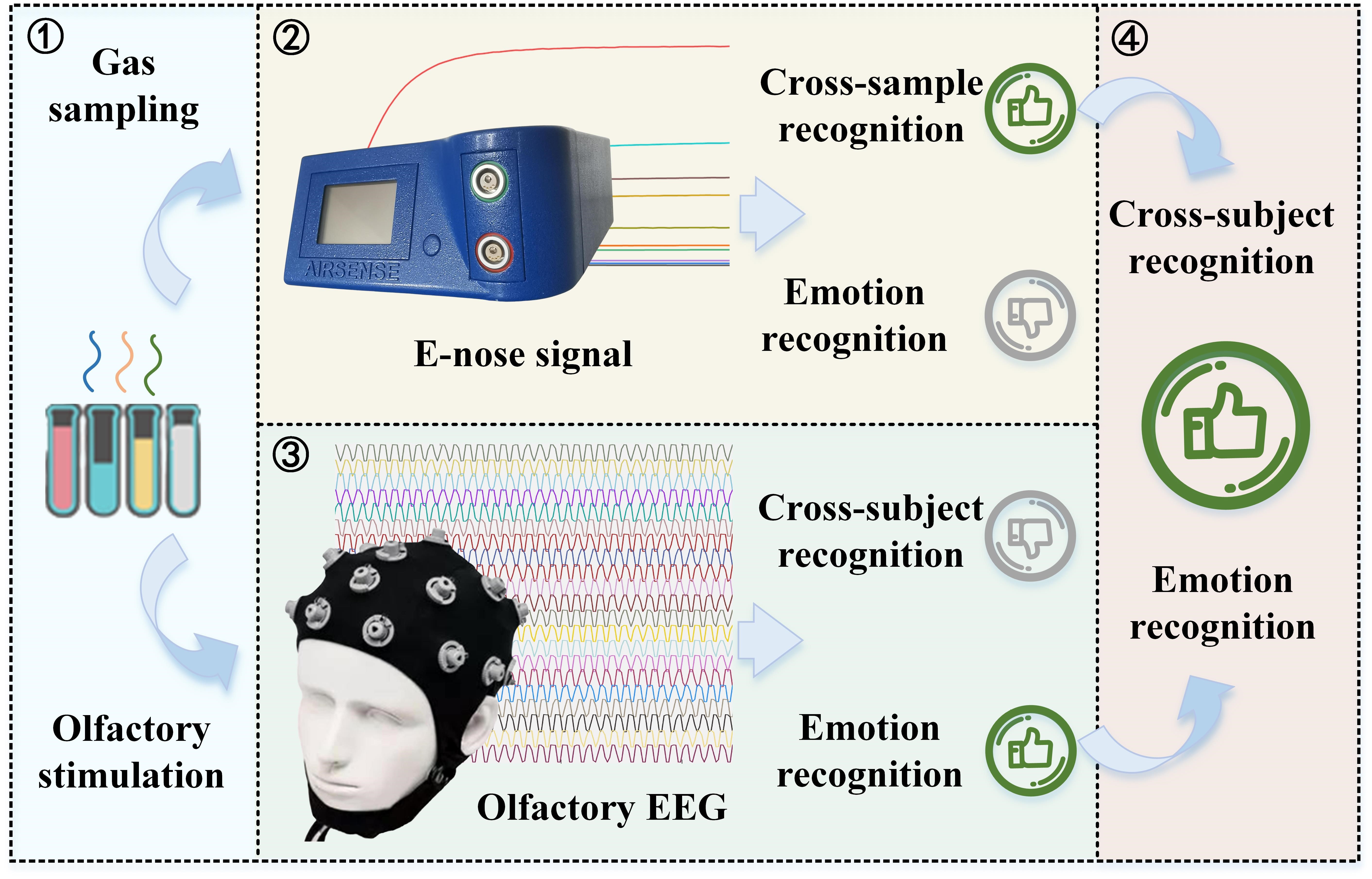}
\caption{Our motivation for fusing olfactory EEG and E-nose signals. 
\label{overflow}}
\end{figure}

\vspace{-0.3cm}
In recent years, the development of multimodal learning has inspired us to solve the olfactory EEG and E-nose signals recognition problem. Usually, different signals based on the same task are complementary, representing the relevant features of the task from different aspects. He et al. used the EEG signal with high temporal and low spatial resolution and the functional near-infrared spectroscopy signal with low temporal and high spatial resolution for motor imagery decoding to complement their temporal and spatial features [18]. Liu et al. combined facial expression with visual EEG and used the cognitive and visual domains to establish a multimodal coupling model to recognize facial emotion [10]. In this paper, an olfactory EEG and E-nose multimodal learning method for cross-subject olfactory preference recognition is proposed. Different from the monomodal recognition of the E-nose or olfactory EEG, the core idea of the proposed method is to complement the cross-sample recognition ability of the E-nose and the emotion recognition ability of olfactory EEG (Fig. 1: \circled{1}+\circled{2}+\circled{3}+\circled{4}). It should be emphasized that the E-nose contains the feature information that only represents odors, while the olfactory EEG is influenced by both odors and individual differences. Specifically, the proposed method mines the common features between the olfactory EEG and E-nose for representing odor information. In addition, E-nose features are used to optimize the spatial distribution of olfactory EEG features, and the optimized olfactory EEG features are used to exploit the individual features that represent the subject’s olfactory preference. Ultimately, the model fuses common features that ensure cross-subject recognition ability and individual features that ensure olfactory preference recognition ability to comprehensively and accurately represent each person’s olfactory preference.
\\\indent Overall, the proposed multimodal learning method of olfactory EEG and E-nose for cross-subject olfactory preference recognition has the following four contributions.
	\\\indent1) An acquisition and preprocessing paradigm for olfactory EEG and E-nose multimodal data is established.
	\\\indent2) A novel strategy for complementary olfactory EEG and E-nose recognition abilities is proposed to recognize cross-subject olfactory preference.
	\\\indent3) The proposed method effectively mines the common features containing odor information between the olfactory EEG and E-nose signals while extracting the individual features in the olfactory EEG that represent the subject’s olfactory preference.
	\\\indent4) Finally, cross-subject olfactory preference recognition is achieved within 24 subjects by fusing the extracted common and individual features, which outperformed state-of-the-art recognition methods. In addition, the unique advantages of the proposed method for cross-subject olfactory preference analysis provide technical support for the practical application of odor evaluation.

\section{Related work}
Our work mainly involves E-nose and olfactory EEG signal recognition and multimodal learning. Recent advances in these areas are briefly reviewed in this section.

\subsection{E-nose signal recognition}

E-nose technology has proven to be a valuable method for odor evaluation. It has been applied in critical areas such as food analysis and quality identification [19], [20]. Compared with traditional olfactory sensory evaluation methods, E-nose has the advantages of high sensitivity, reliability, and rapidity while ensuring high bionic. Previously, traditional machine learning methods have been widely applied to E-nose recognition. The main methods are k-nearest neighbor classifier, extreme learning machine, linear discriminant analysis, support vector machine, and principal component analysis [21], [22]. However, many classical identification algorithms have fixed model frameworks and fewer adjustable parameters, which limits their generalization ability. In addition, such principal component analysis methods usually require complex feature engineering, which limits their applications. In recent years, deep learning has been widely used in image recognition, natural language processing, fault diagnosis, and other fields. It has also profoundly influenced the innovation of E-nose recognition technology based on its powerful representation and adaptive ability. Peng et al. earlier applied CNN to E-nose recognition, and they designed a deep CNN with up to 38 layers to recognize four different odors [23]. Feng et al. proposed an augmented CNN that better compensates for the existing models' bias to solve the sensor drift problem [24]. Zhao et al. proposed a one-dimensional deep CNN with multi-label capability to comprehensively extract and classify the features of gas mixtures [25]. In short, traditional machine learning methods require less computation and have generally advantages in small sample E-nose signal recognition. However, in most cases, deep learning methods have higher recognition accuracy and stronger robustness. In addition, higher complexity models do not necessarily have better recognition results for low-time complexity E-nose signals. The constructed lighter and more efficient models have become mainstream.

\subsection{Olfactory EEG recognition}
Initially, many studies mainly used traditional machine learning methods to recognize olfactory EEG. They usually extracted the features of olfactory EEG in time and frequency domains and then input the most favorable features into a classifier for odor or emotion recognition. For example, Xia et al. divided olfactory EEG into five frequency bands according to physiological rhythms and then constructed a functional brain network using mutual information. Finally, the network properties were extracted and inputted into a support vector machine classifier for odor and pleasantness recognition [13]. Ezzatdoost et al. extracted the nonlinear and chaotic features of olfactory EEG and inputted them into a linear discriminant analysis classifier to classify four odors [26]. Aydemir et al. used the k-nearest neighbor algorithm to classify the autoregressive model parametric features of the olfactory EEG signals, ultimately identifying the odors [27]. However, the decoding ability of traditional machine learning methods is insufficient for olfactory EEG signals with high temporal resolution and complexity. Deep learning models represented by CNN have been gradually used for olfactory EEG recognition instead of traditional machine learning due to their powerful decoding capabilities. Although deep learning methods effectively recognize olfactory EEG, the models often suffer from overfitting, mainly manifesting poor cross-subject recognition ability.

\subsection{Multimodal learning}
Multimodal learning inspires us to solve olfactory EEG recognition models’ poor cross-subject recognition ability. Real-world information comes from multiple modalities, each portraying the world from its perspective. The information they portray has both similarities and differences. Multimodal learning methods attempt to comprehensively represent the real world by finding a joint representation between multiple modalities. According to this idea, information is greatly interacted between the fields of image recognition, natural language processing, and speech recognition [28], [29], [30]. Such text generation, image description, and speech localization tasks are well realized by establishing correspondences between images, text, or audio. In the field of physiological signal decoding, EEG, facial expression [10], functional near-infrared spectroscopy [18], and electrooculogram [31] are becoming increasingly linked. Usually, the multimodal features are complementary, and the interaction between multimodal features allows the model to learn the relevant task information from different perspectives, giving the model a greater generalization ability. This work draws on multimodal complementarity and represents individual's olfactory preference through multimodal signals from olfactory EEG and E-nose. Among them, the common features (containing odor information) between the olfactory EEG and E-nose ensure the model’s cross-subject recognition ability. The olfactory EEG's individual features (containing emotional information) ensure the model’s olfactory preferences representation ability.

\section{Experiments and methods}
\subsection{Experimental}
\subsubsection{Instrument}  \indent

\begin{figure}[t]
\centering
\includegraphics[width=90mm]{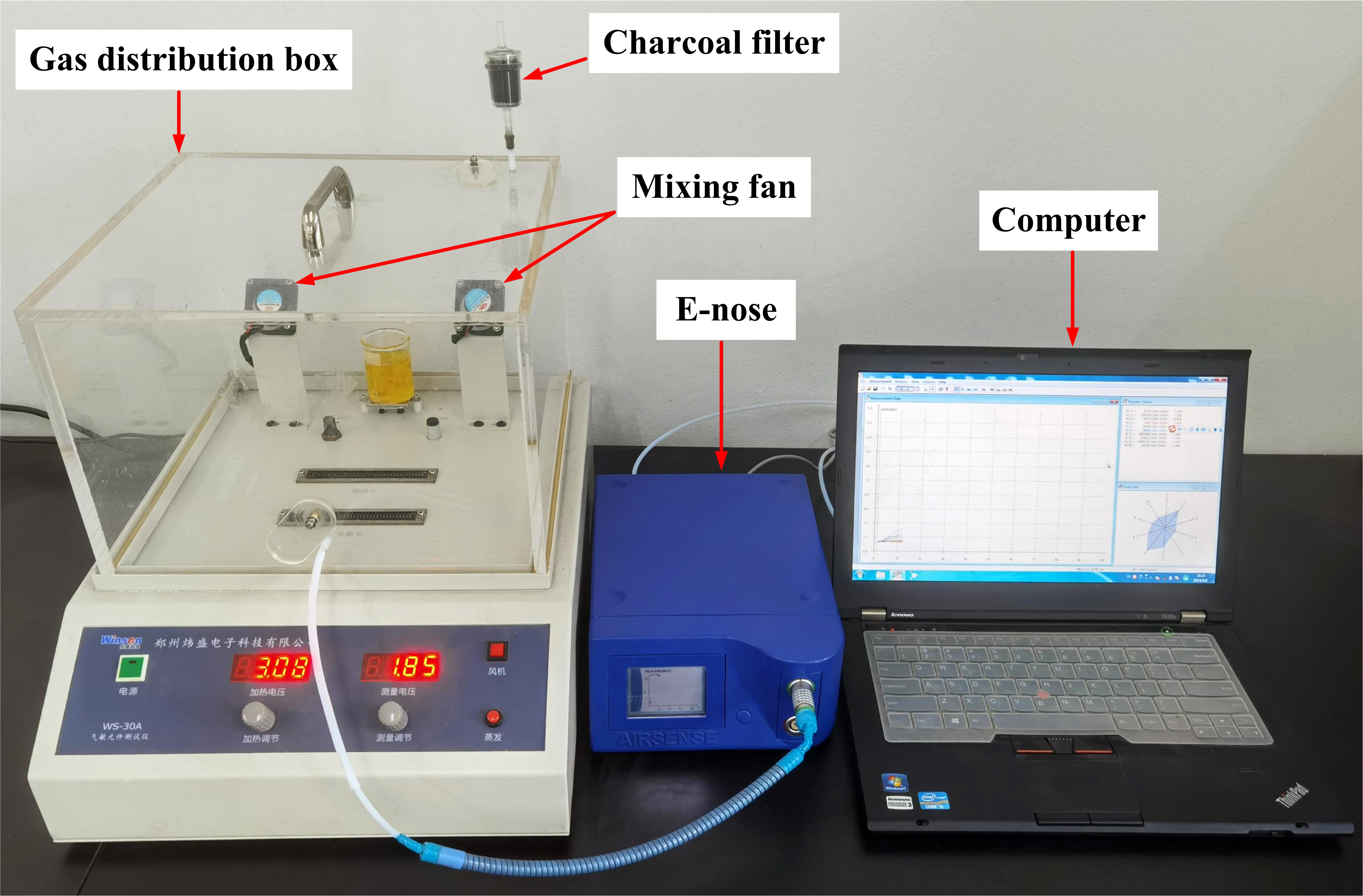}
\caption{E-nose sampling system. 
\label{overflow}}
\end{figure}

The structure of the E-nose sampling system is shown in Fig. 2, which consists of a WS-30A gas-sensitive component test system (Zhengzhou Weisheng Electronic Technology Co., Ltd., China), PEN3 E-nose (AirSense Analytics Inc. Schwerin, Germany), and Thinkpad computer (Lenovo Group Ltd., China). The WS-30A system mainly comprises a gas distribution box, a charcoal filter, and a mixing fan. The E-nose integrates a gas sensor array, a signal processing unit, and a pattern recognition system. There are 10 metal oxide sensors in the gas sensor array. Table I shows the E-nose sensors' main performance. Different sensors have different sensitivities to gases. The sensor and the measured gas undergo a redox reaction during contact, and the sensor resistance decreases or increases gradually with the partial pressure of the gas. The gas information is detected by the conductivity value $G/G_0$ ($G$ is the conductivity value when the sensor is in contact with the sample, and $G_0$ is when the sensor is in contact with the activated carbon filtered air). During the E-nose signal collection, the outside gas enters the gas distribution box after being filtered by the charcoal filter, and the gas inside the gas distribution box is evenly distributed through the mixing fan. Then the gas enters the E-nose and comes into complete contact with the gas sensors. And the conductivity values of the sensors are collected through the signal processing unit. Finally, the collected signals are transmitted to the computer and analyzed by the pattern recognition system.

\begin{table}[t]
\caption{THE MAIN PERFORMANCE OF E-NOSE SENSORS}
\begin{center}
\begin{tabular}{ccccc}
\hline
NO. & Sensor & Main performance                                                                         & \begin{tabular}[c]{@{}c@{}}Selective\\    \\ gas\end{tabular} & \begin{tabular}[c]{@{}c@{}}Detectability\\    \\ (ppm)\end{tabular} \\ \hline
1   & W1C    & Aromatics                                                                                & Toluene                                                       & 10                                                                  \\
2   & W5S    & Nitride   oxides                                                                         & NO2                                                           & 1                                                                   \\
3   & W3C    & Ammonia,   aroma constituent                                                             & Benzene                                                       & 10                                                                  \\
4   & W6S    & Hydrogen                                                                                 & H2                                                            & 100                                                                 \\
5   & W5C    & Alkenes,   aroma constituent                                                             & Propane                                                       & 1                                                                   \\
6   & W1S    & Broad-methane                                                                            & CH4                                                           & 100                                                                 \\
7   & W1W    & Sulfur-containing   organics                                                             & H2S                                                           & 1                                                                   \\
8   & W2S    & Broad alcohols                                                                           & CO                                                            & 100                                                                 \\
9   & W2W    & \begin{tabular}[c]{@{}c@{}}Aroma   constituent, sulfur \\ organic compounds\end{tabular} & H2S                                                           & 1                                                                   \\
10  & W3S    & Methane   and aliphatic                                                                  & CH3                                                           & 10                                                                  \\ \hline
\end{tabular}
\end{center}
\end{table}

Olfactory EEG evoked device. The self-developed low-noise, high-stability olfactory EEG evocation system consists of an air generator, the WS-30A system, a gas diffusion module, and a control module. Among them, the air inlet of the WS-30A’s gas distribution box is connected to the air generator, and the air outlet is connected to the gas diffusion module. Finally, the experimental odor is delivered to the front of the subject’s nose through the gas diffusion module. During this period, the WS-30A’s mixing fan always operates at constant power to ensure uniform gas distribution. The control module achieves timed and quantitative olfactory stimulation.
\\\indent Olfactory EEG acquisition system. The olfactory EEG signals are acquired by an NCERP-P EEG acquisition system (Shanghai NCC Electronics Co., Ltd., China) with a 256 Hz sampling frequency. And 21 saline conductive electrodes (Fz, Cz, Pz, T3, T4, C3, C4, Fp1, Fp2, F7, F8, T5, T6, O1, O2, F3, F4, P3, P4, A1, A2) of the EEG caps (GreentekPty. Ltd., China) are arranged according to the 10-20 system.

\subsubsection{Subjects and Materials}  \indent
\\\indent The study protocol followed the revised Declaration of Helsinki. It was approved by the Scientific Research Ethics and Science and Technology Safety Committee of Northeastern Electric Power University. 24 right-handed subjects (14 males and 10 females, aged 22 to 26 years) were recruited to participate in this study. They signed an informed consent form and were informed about the details of the experiment. They had not suffered from a recent cold and were also free of olfactory or psychiatric disorders.
\\\indent In this study, the odors evolved from gardenia perfume, amaranth fermented brine, grated puree of houttuynia cordata, and peach jam left to stand for 3 minutes were used to evoke olfactory EEG and E-nose signals. The details of the experimental materials are shown in Table II.

\begin{table}[t]
\caption{INFORMATION OF FOUR KINDS OF EXPERIMENTAL MATERIALS}
\begin{center}
\begin{tabular}{ccccc}
\hline
NO. & Odors  & Composition                                             \\
\hline
1   & Aroma  & 0.1 ml gardenia perfume,   30 ml distilled water        \\
2   & Stink  & 20 ml amaranth fermented brine, 30 ml   distilled water \\
3   & Fishy  & 10 g houttuynia cordata, 30 ml distilled   water        \\
4   & Fruity & 20 g peach jam, 30 ml distilled water            \\ \hline  
\end{tabular}
\end{center}
\end{table}

\begin{figure}[t]
\centering
\includegraphics[width=90mm]{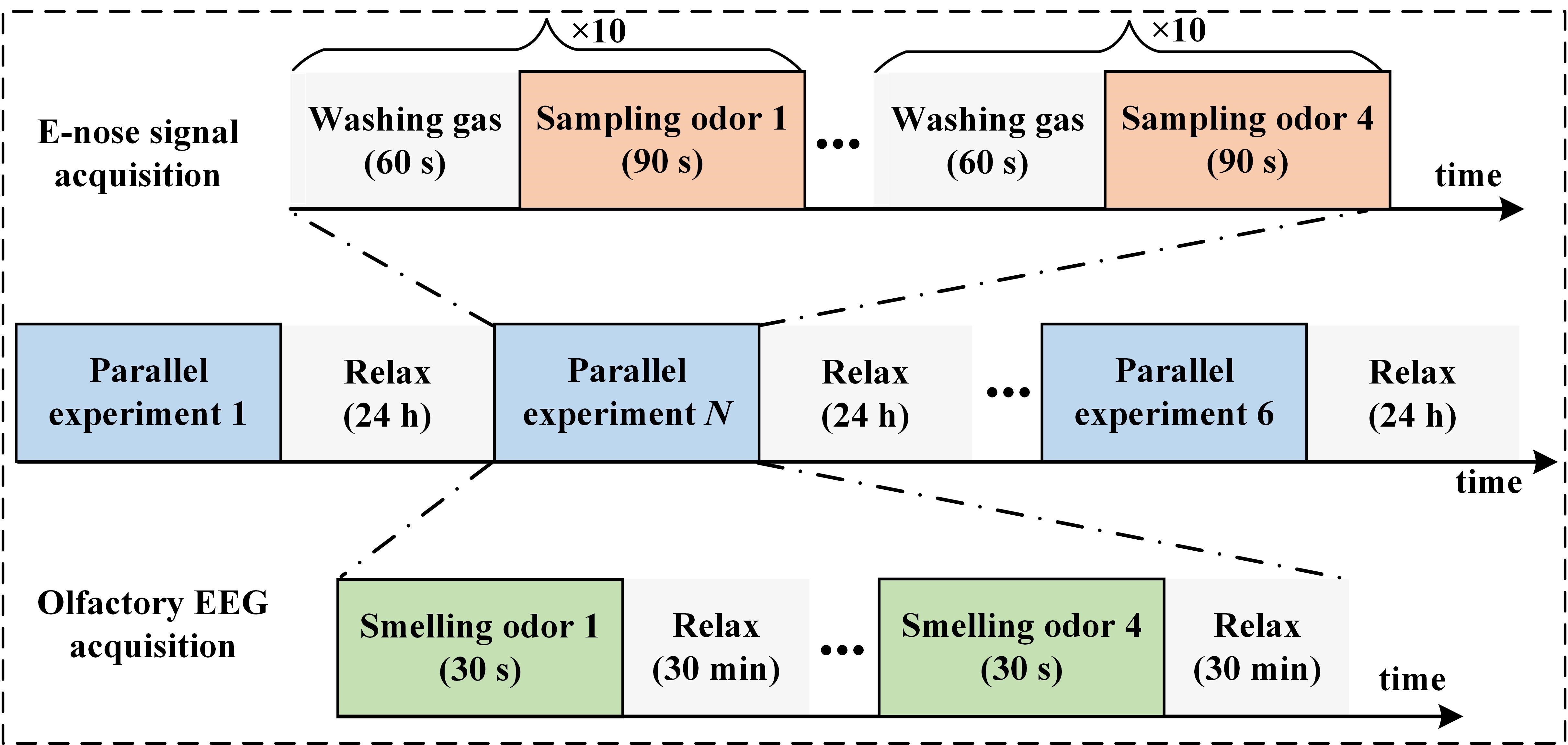}
\caption{Olfactory EEG and E-nose signals acquisition process. 
\label{overflow}}
\end{figure}

\subsubsection{Data Acquisition}  \indent
\\\indent The olfactory EEG and E-nose signals acquisition process is shown in Fig. 3. For olfactory EEG acquisition, each subject participates in six parallel experiments, and the interval between each parallel experiment was 24 hours. In each parallel experiment, subjects’ olfactory EEG under four odor stimuli were collected separately for 30 seconds, in which the subjects rested for 30 minutes after each olfactory EEG acquisition. E-nose signal acquisition was performed in parallel with the olfactory EEG acquisition to establish the same number of E-nose samples as the olfactory EEG samples for each subject. In each E-nose signal acquisition parallel experiment, 10 sets of E-nose signals were acquired for 90 seconds each under each odor stimulus. In addition, the gas sensor array was washed for 60 seconds before each E-nose signal was acquired to return the sensor to the baseline state.

\begin{figure*}[t]
\centering
\includegraphics[width=183mm]{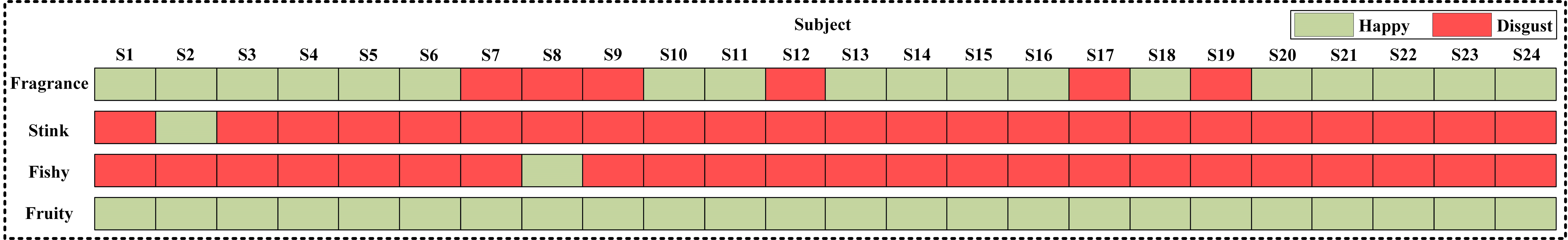}
\caption{Evaluation of subjects’ preference for experimental odors. 
\label{overflow}}
\end{figure*}

\subsubsection{Preprocessing and Experimental Setup}  \indent
\\\indent In this experiment, 24 subjects’ olfactory EEG under four odor stimuli were acquired. Each odor stimulus generates six 30-second pieces of parallel data. 1 - 21 seconds of each parallel data were retained. The size of each olfactory EEG sample was 2 seconds. Then, they were band-pass filtered from 0.5-45 Hz, and the filtered olfactory EEG samples with a sampling frequency of 256 Hz were downsampled to 128 Hz to reduce the data amount. Eventually, 5760 (24 × 4 × 6 × 10) olfactory EEG samples were created. As shown in Fig. 4, the labels of the 5760 olfactory EEG samples were set up according to the subject’s preference. For the E-nose, the sampling frequency was 1 second, and the size of each E-nose sample was 90 seconds. Consequently, 5760 E-nose samples were established. Finally, by combining the 5760 EEG samples one-to-one with the E-nose samples collected in parallel during the experiment, 5760 multimodal samples were created with labels matching those of the olfactory EEG portion.

\begin{figure*}[t]
\centering
\includegraphics[width=183mm]{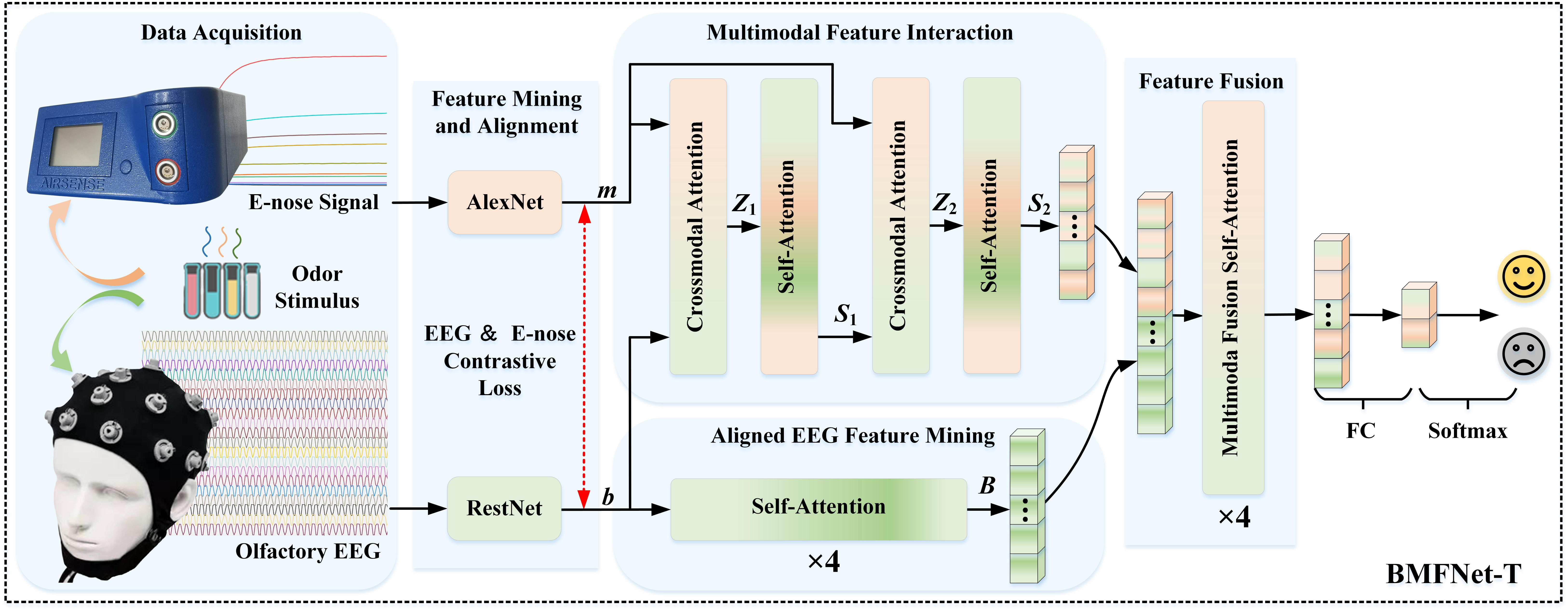}
\caption{Overview of the proposed BMFNet-T.
\label{overflow}}
\end{figure*}

\subsection{Olfactory EEG and E-nose Multimodal Learning Method}

Human olfactory preference is influenced by odor and individual differences. The proposed method mines the common features containing odor information between the olfactory EEG and E-nose while extracting the individual features in the olfactory EEG that represent the subject’s olfactory preference. Ultimately, the common and individual features are combined to represent a person’s odor preferences comprehensively. The proposed BMFNet teacher (BMFNet-T) model is shown in Fig. 5. It mainly contains a feature mining and alignment (FMA) module, a multimodal feature interaction (MFI) module, an aligned EEG feature mining (AEFM) module, and a feature fusion (FF) module. The detailed process of the proposed method is as follows. Firstly, AlexNet [32] and RestNet [33] are used to extract the initial features of E-nose and olfactory EEG signals, respectively. Secondly, the olfactory EEG and E-nose initial features are aligned using contrastive loss. Thirdly, the aligned olfactory EEG and E-nose features are used for feature interaction to exploit their commonalities fully. Meanwhile, the aligned olfactory EEG features are used to extract the individual features in olfactory EEG. Finally, the multimodal common features are fused with the olfactory EEG individual features to recognize cross-subject olfactory preference.

\subsubsection{Feature Mining and Alignment}  \indent
\\\indent The olfactory EEG signal is more complex than the E-nose signal, which determines that the deeper network is more suitable for mining olfactory EEG features. So, the shallow AlexNet and the deeper RestNet extract initial features for E-nose and olfactory EEG signals, respectively. In this process, the input multimodal samples of the E-nose and olfactory EEG are reshaped to (1, 10, 90) and (1, 21, 256), respectively. Then, the convolution kernel size, step size, padding size, output channels, pooling kernel size, and pooling kernel step size are adjusted to make the extracted multimodal initial features of the same size. And the output monomodal features of AlexNet and RestNet with dimensions (64, 1, 5) and (160, 1, 2) are reshaped to (1, 16, 20), respectively. Finally, initial features of the olfactory EEG and E-nose signals are obtained but unaligned. To better fuse the multimodalities features, the initial features are aligned through mean squared error loss [34].

\subsubsection{Common and Individual Features Mining}  \indent
\\\indent The aligned initial features $m$ and $b$ are input into the MFI module to mine the common features between olfactory EEG and E-nose signals. The MFI module mainly comprises crossmodal attention and self-attention modules, which are alternately connected. The first crossmodal attention module’s output $Z_1$ is the first self-attention module’s input. Then, the first self-attention module’s output $S_1$ and the initial features $m$ are used as the input of the second crossmodal attention module. Finally, the second crossmodal attention module’s output $Z_2$ is input into the second self-attention module to get the final output $S_2$.
\indent The structure of the crossmodal attention module is shown in Fig. 6 (a). It mainly comprises Embedding, layer normalization (LN), 
Linear, multi-head attention (MHA), and multi-layer perceptron (MLP). Firstly, $m$ and $b$ with 1 × 16 × 20 size are reconstructed into 2D sequences by 2D convolution. The kernel size and step size of the 2D convolution are 1 × 2. The reconstructed sequence contains 160 tokens to ensure each feature information is reconstructed efficiently. The output channel of the 2D convolution is set to 320 to ensure that valid information is retained. Then, classification tokens are added before the reconstructed sequence for classification. So after the Embedding, the monomodal features in data form of 161 × 320 are obtained. Secondly, the monomodal features are normalized by LN to ensure the stability of feature distribution. In the Linear, the normalized features are multiplied with the optimizable matrix to obtain the inputs $Q_m$, $K_b$, and $V_b$ of MHA. Where $Q_m$ is the query value to find the common features of the olfactory EEG and E-nose signals, $K_b$ represents the key feature information in the olfactory EEG, and $V_b$ represents all the features in the olfactory EEG. Thirdly, $Q_m$, $K_b$, and $V_b$ are fed into the MHA to mine the common features between the olfactory EEG and E-nose signals. To preserve the underlying features of the E-nose, the output of the MHA is summed with $m$ to obtain $Y$. Then, $Y$ passes through the LN and the MLP, where the MLP consists of two fully connected layers (FC), and the GeLU activation function is inserted into the FC layers to make the model training process more robust. Finally, the output of the MLP is summed with $Y$ to obtain the output of the crossmodal attention module $Z$. It is computed as follows.

\begin{equation}
Y = {\mathop{\rm MHA}\nolimits} ({Q_m},{K_b},{V_b}) + m
\end{equation}

\begin{equation}
Z = {\mathop{\rm MLP}\nolimits} ({\mathop{\rm LN}\nolimits} (Y)) + Y
\end{equation}

\begin{figure}[t]
\centering
\includegraphics[width=90mm]{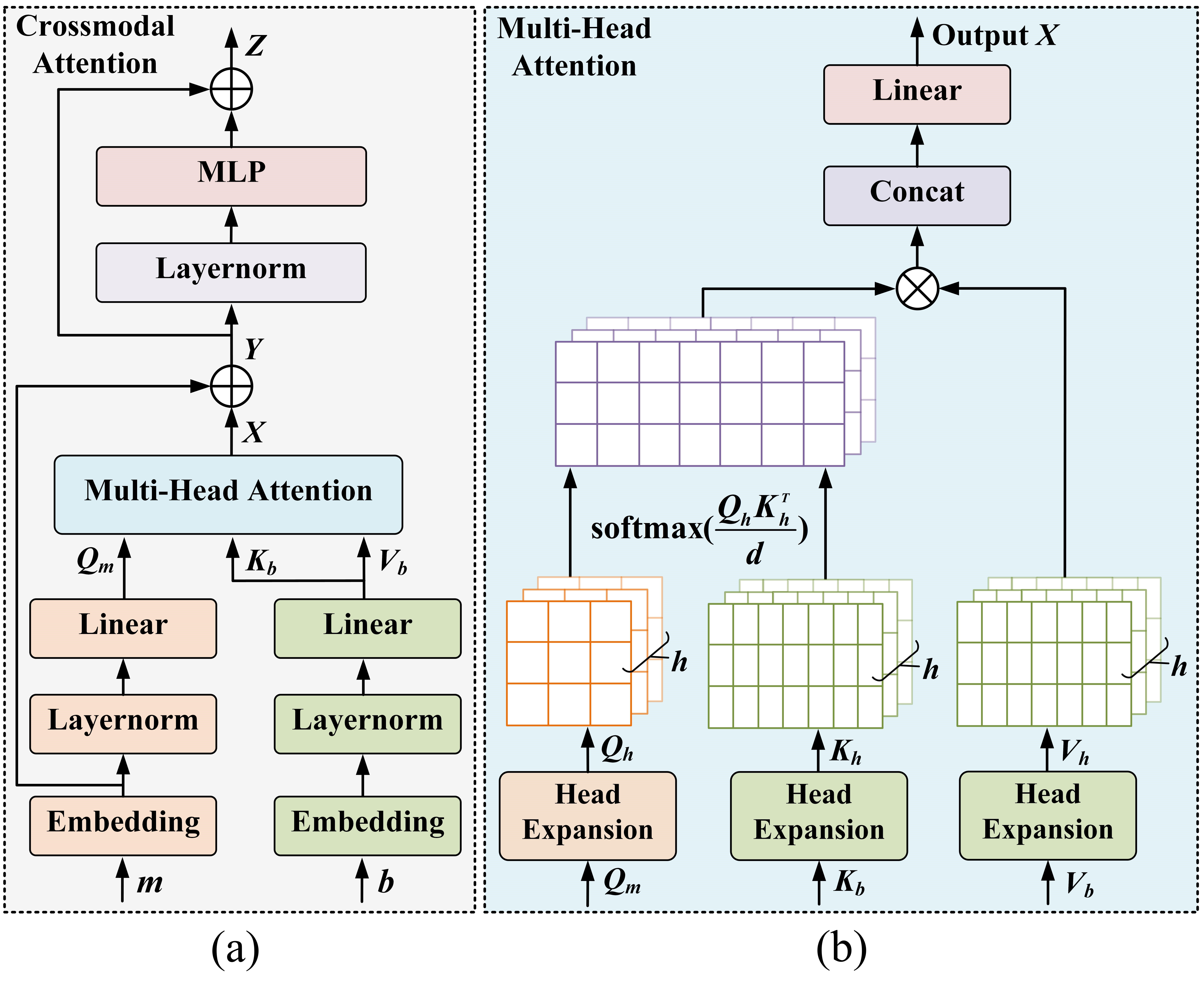}
\caption{(a) Architectural elements of crossmodal attention module and (b) calculation flow of MHA.
\label{overflow}}
\end{figure}

MHA is a key part of the crossmodal attention module for fully exploiting multimodal features. Compared to crossmodal attention with one head, the multi-head in MHA allows the model to fully mine the common between complex subspaces features from different modalities. The calculation flow of MHA is shown in Figure 6 (b). Firstly, the input $Q_m$, $K_b$, and $V_b$ are expanded from 161 × 320 to 161 × 8 × 40 along the second dimension. The first and second dimensions of the expanded features are interchanged to obtain the multi-head features $Q_h$, $K_h$, and $V_h$, where the number of heads is 8. Secondly, the second and third dimensions of $K_h$ are interchanged. After the interchanging, $K_h$ is multiplied by $Q_h$ and scaled by dividing $d$. Thirdly, the scaled values are input to the softmax layer and multiplied with $V_h$ to obtain a multi-head feature. Finally, the first and third dimensions of the multi-head feature are connected to integrate the multi-head subfeature. The integrated feature is multiplied by an optimizable matrix $W_h$ to obtain the linearized output $X$. Its calculation formula follows.

\begin{equation}
X = {\mathop{\rm Concat}\nolimits} ({\mathop{\rm softmax}\nolimits} (\frac{{{Q_h} \times {K_h}^T}}{d}) \times {V_h}) \times {W_h}
\end{equation}

\noindent Where $T$ represents the second and third dimensions are interchanged, d = ${\sqrt {40}}$, Concat represents merging the first and third dimensions of the feature matrix.

The self-attention module is similar to the crossmodal attention module, where the Embedding, LN, Linear, MHA, and MLP have the same structure. The difference is the input of the self-attention module is a monomodal feature matrix, and only a single Embedding, LN, and Linear are connected in series in turn.

The AEFM module is used to mine individual features of human odor preferences in olfactory EEG, which mainly consists of four self-attention modules connected in series. The structure of the self-attention module is the same as the self-attention module in the MFI module. As a result, the size of the output feature map $B$ is 161 × 320.

\subsubsection{Multimodal Feature Fusion}  \indent
\\\indent For multimodal feature fusion, the classification token of feature sequences $S_2$ and $B$ are spliced into a 1 × 640 feature sequence and reshaped into a 1 × 16 × 40 feature matrix as the input of the FF module. The FF module consists of four multimodal fusion self-attention modules connected in series. Its self-attention module is similar to the self-attention module in the AEFM module, only the Embedding module is slightly different. The Embedding module has a kernel size and step size of 2 × 4 for the 2D convolution, and the output channel is set to 320, so 80 tokens can be obtained in the reconstructed sequence. Then, the feature sequence of size 81 × 320 is obtained by adding classification tokens before the reconstructed sequence. Thus, after four multimodal fusion self-attention modules connected in series, their classification tokens of size 1 × 320 are fed into the FC layer. Finally, the olfactory preference prediction is obtained through the softmax layer.

\begin{figure}[t]
\centering
\includegraphics[width=90mm]{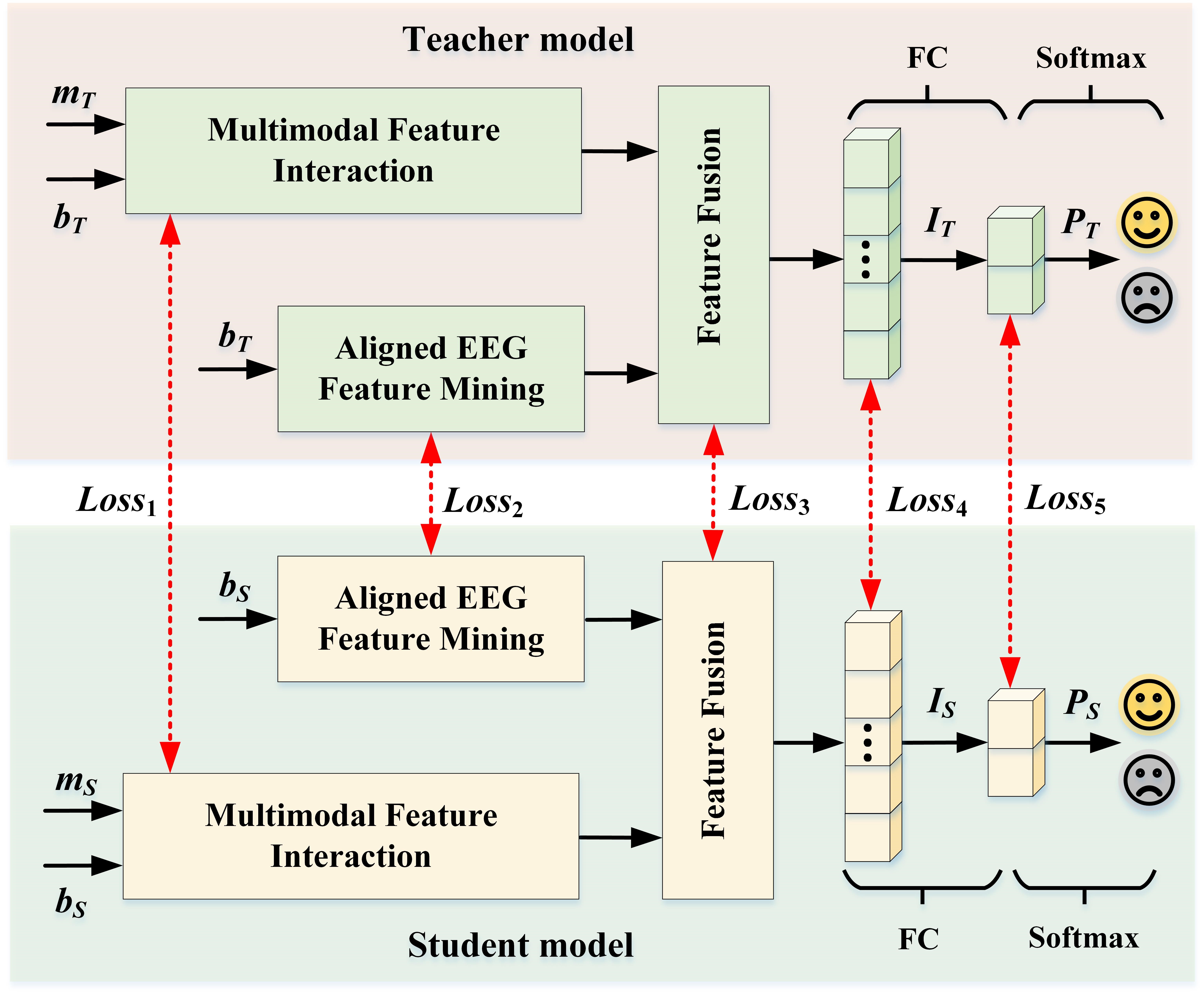}
\caption{Proposed knowledge distillation framework.
\label{overflow}}
\end{figure}

\subsubsection{Multimodal Knowledge Distillation}  \indent
\\\indent Knowledge distillation is a domain adaptation technique that transfers knowledge from the teacher to the student model. The student model maintains comparable recognition capabilities to the teacher model while having fewer parameters and computational quantity. This work uses knowledge distillation to refine the multimodal knowledge of the BMFNet-T model. The BMFNet-S model also has an FMA module, an MFI module, an AEFM module, and an FF module. The FMA modules in the BMFNet-S and BMFNet-T models are the same. In contrast to the BMFNet-T model, the BMFNet-S model has only one crossmodal attention module and a self-attention module in the FMA module. The AEFM module has only two self-attention modules in a series. The FF module has only two multimodal fusion self-attention modules in a series. The framework of knowledge distillation is shown in Fig. 7. The knowledge in the BMFNet-T model is refined into the BMFNet-T model by optimizing the loss functions $Loss_1$, $Loss_2$, $Loss_3$, $Loss_4$, and $Loss_5$. Among them, for the MFI module, the 2nd crossmodal attention module and the 2nd self-attention module of the BMFNet-T model are used to guide the crossmodal attention module and the self-attention module of the BMFNet-S model, respectively. For the AEFM module, the BMFNet-T model’s 2nd and 4th self-attention modules are used to guide the BMFNet-S model’s 1st and 2nd self-attention modules, respectively. For the FF module, the BMFNet-T model’s 2nd and 4th multimodal fusion self-attention modules are used to guide the BMFNet-S model’s 1st and 2nd multimodal fusion self-attention, respectively. In addition, the BMFNet-T model’s FC layer and softmax layer are used to guide the BMFNet-S model’s FC layer and softmax layer, respectively.

\begin{figure}[t]
\centering
\includegraphics[width=90mm]{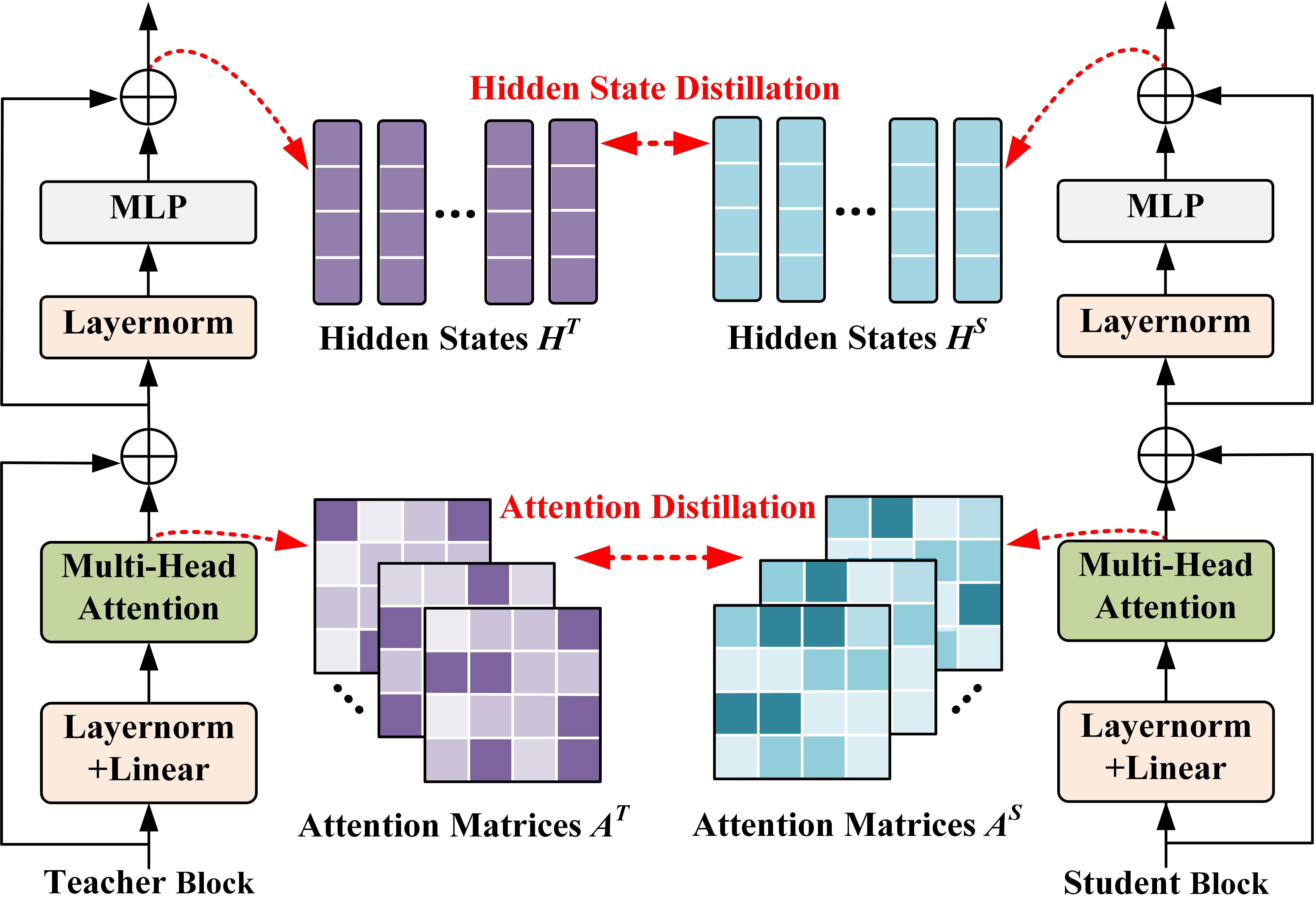}
\caption{The illustrations of transformer block distillation consist of attention loss and hidden loss.
\label{overflow}}
\end{figure}

The transformer block distillation is used in the knowledge distillation of the MFI, AEFM, and FF modules. Its details are shown in Fig. 8. Transformer block distillation includes attention distillation and hidden state distillation. The attentional distillation can better capture knowledge representing the relationships between multimodal features. The definitions are as follows.

\begin{equation}
{L_{{\rm{attn}}}} = \frac{1}{h}\sum\nolimits_{i = 1}^h {{\mathop{\rm MSE}\nolimits} } (A_i^T,A_i^S)
\end{equation}
\noindent  Where $h$ is the number of attention heads, ${A_i} \in {{\mathbb R}^{l{\rm{ \times }}l}}$ refers to the attention matrix corresponding to the $i$th head in the teacher or student block, and $l$ is the length of the input sequence. The hidden state distillation is used to distill multimodal knowledge fully, which is defined as follows.

\begin{equation}
{L_{hidn}} = {\mathop{\rm MSE}\nolimits} ({H^T},{H^S})
\end{equation}
\noindent Where the matrices ${H^T} \in {{\mathbb R}^{l \times e}}$ and ${H^S} \in {{\mathbb R}^{l \times e}}$denote the hidden states in the teacher block and student block, respectively. The scalar value $e$ denotes the hidden size of the teacher and BMFNet-S models.

$Loss_1$, $Loss_2$, and $Loss_3$ are expressed as follows.
\begin{equation}
Los{s_j} = {{\mathop{\rm Module}\nolimits} _j}(\sum\nolimits_{i = 1}^n {{L_{attn}}}  + \sum\nolimits_{i = 1}^n {{L_{hidn}}} )
\end{equation}
\noindent Where $j$ = 1, 2, 3. Module1, Module2, and Module3 denote the transformer blocks used for distillation in the MFI, AEFM, and FF modules, respectively. $n$ is the number of transformer blocks in each distill module.

\begin{table}[htbp]
  \centering
    \begin{tabular}{p{27em}}
    \toprule
    \textbf{Algorithm 1} Knowledge distillation procedure \\
    \midrule
    \textbf{Input:} Training dataset $X$ = {$x_1, x_2, ···,x_n$}, initialize the  
    network parameters ${\theta _t}$ and ${\theta _s}$ of BMFNet-T and 
    BMFNet-S, respectively. \\
    \textbf{Hyperparameters:} $Batch$, $Epochs$, $N\_batch_j$=$n$/Batch \\
    \textbf{1: for}    each batch sequence pair $X_j \in X$ \textbf{do} \\
    \textbf{2:\quad }       $X$ = { $X_1$, $X_2$, ···,$X_N\_batch$ } \\
    \textbf{3:   end} \\
    \textbf{4:}   /* Training BMFNet-T model */ \\
    \textbf{5: for} $i$ = 1, 2, ···, $Epochs$ \textbf{do} \\
    \textbf{6: \quad   for} $j$ = 1, 2, ···, $N\_batch$  \textbf{do} \\
    \textbf{7: \qquad}         loss(${\theta _t}$) ← BMFNet-T($X_j$, ${\theta _t}$) \\
    \textbf{8: \qquad}         update ${\theta _t}$ \\
    \textbf{9:  \quad   end} \\
    \textbf{10:  end} \\
    \textbf{11:}  /* Training BMFNet-S model */ \\
    \textbf{12:  for} $i$ = 1, 2, ···, $Epochs$ \textbf{do} \\
    \textbf{13:  \quad   for} $j$ = 1, 2, ···, $N\_batch$ \textbf{do} \\
    \textbf{14:}  \qquad   loss(${\theta _s}$) ← BMFNet-S($X_j$, ${\theta _s}$) \\
    \textbf{15:}   \qquad   update ${\theta _s}$ \\
    \textbf{16:   \quad    end} \\
    \textbf{17:  \quad for} $j$ = 1, 2, ···, $N\_batch$ \textbf{do} \\
    \textbf{18:}  \qquad        $W_t$ ← BMFNet-T($X_j$, ${\theta _t}$) \\
    \textbf{19:}  \qquad    loss(${\theta _s}$), $W_s$ ← BMFNet-S($X_j$, ${\theta _s}$) \\
    \textbf{20:}  \qquad   loss(${\theta _s}$) = loss(${\theta _s}$) + loss($W_t$, $W_s$) \\
    \textbf{21:}  \qquad   update ${\theta _s}$ \\
    \textbf{22: \quad end} \\
    \textbf{23:  end} \\
    \bottomrule
    \end{tabular}%
  \label{tab:addlabel}%
\end{table}%

\indent Algorithm 1 describes the process of knowledge distillation. The steps are BMFNet-T model training, BMFNet-S model training, and BMFNet-T model guiding BMFNet-S model training. BMFNet-T model is trained with the following loss function.
\begin{equation}
Los{s_T} = {\mathop{\rm MSE}\nolimits} ({m_T},{b_T}) + {\mathop{\rm CE}\nolimits} ({P_T},label)
\end{equation}
\noindent Where CE(·) denotes the cross entropy between two probability distributions, $m_T$ and $b_T$ are the aligned E-nose and olfactory EEG feature in the BMFNet-T model, respectively.

\indent The loss function for training the BMFNet-S model using hard labels is as follows.
\begin{equation}
Los{s_{hard}} = {\mathop{\rm MSE}\nolimits} ({m_S},{b_S}) + {\mathop{\rm CE}\nolimits} ({P_S},label)
\end{equation}

\noindent Where ${m_S}$ and ${b_S}$ are the aligned E-nose and olfactory EEG feature maps in the BMFNet-S model, respectively. 
    The ${Loss_S}$ is optimized when jointly training the BMFNet-S model using hard labels and soft label losses are as follows.
\begin{equation}
Los{s_S} = \alpha  \times {L_{hard}} + (1 - \alpha ) \times {L_{soft}}
\end{equation}
\noindent Where $\alpha  \in (0,1)$, and it is a hyperparameter to optimize the ${Loss_S}$.

\indent For the FC layer, the definition of knowledge distillation is as follows.
\begin{equation}
Los{s_4} = {\mathop{\rm MSE}\nolimits} ({I_T},{I_S})
\end{equation}
\noindent Where the  ${I_T} \in {{\mathbb R}^e}$ and ${I_S} \in {{\mathbb R}^e}$ denote the FC layer features in the teacher and student models, respectively.

\indent For the softmax layer, the knowledge distillation is defined as follow.
\begin{equation}
Los{s_5} = {\mathop{\rm KL}\nolimits} (\log (\frac{{{P_T}}}{T}),\log (\frac{{{P_S}}}{T})) \times {T^2}
\end{equation}
\noindent Where KL(·) is the Kullback-Leibler divergence loss function, the  ${P_T} \in {{\mathbb R}^c}$ and  ${P_S} \in {{\mathbb R}^c}$ denote the softmax layer features in the teacher and student models, respectively. The $c$ is the number of sample labels, and $T$ is the distillation temperature.

\begin{table}[t]
\caption{THE COMPARISON OF THE BMFNET-T AND BMFNET-S MODELS}
\begin{center}
\begin{tabular}{ccccc}
\hline
Evaluation index & BMFNet-T & BMFNet-S         \\
\hline
Params (M)       & 9.06     & 5.35 (↓3.71)     \\
FLOPs (G)        & 257.74   & 148.39 (↓109.35) \\
Accuracy (\%)    & 92.88    & 92.79 (↓0.09)    \\
F1-score (\%)    & 92.71    & 92.64 (↓0.07)    \\
Recall (\%)      & 92.93    & 93.11 (↑0.18)    \\
Precision (\%)   & 92.55    & 92.42 (↓0.13)   \\
\hline
\end{tabular}
\end{center}
\end{table}

\begin{table*}[t]
\caption{RECOGNITION RESULTS OF THE STATE-OF-THE-ART MONOMODAL RECOGNITION METHOD}
\begin{tabular}{ccccccccccc}
\hline
               &               &               &             &                &  &  &               &               &             &                \\ \cline{1-5} \cline{7-11} 
Method         & \multicolumn{4}{c}{EEG}                                      &  & \multicolumn{5}{c}{E-nose}                                      \\ \cline{1-5} \cline{7-11} 
               & Accuracy (\%) & F1-score (\%) & Recall (\%) & Precision (\%) &  &  & Accuracy (\%) & F1-score (\%) & Recall (\%) & Precision (\%) \\
AlexNet        & 54.24         & 68.74         & 96.94       & 54.26          &  &  & 90.28         & 90.9          & 89.11       & 92.95          \\
EEGNet         & 54.05         & 64.62         & 89.83       & 52.15          &  &  & 78.3          & 72.28         & 69.93       & 84.4           \\
VGG11          & 54.17         & 69.53         & 97.14       & 54.17          &  &  & 88.56         & 88.52         & 87.26       & 93.84          \\
RestNet18      & 64.29         & 64.85         & 72.74       & 63.22          &  &  & 89.03         & 91.21         & 94.87       & 89.46          \\
DenseNet       & 64.13         & 64.4          & 74.76       & 61.34          &  &  & 83.56         & 87.8          & 97.09       & 82.34          \\
ShuffleNetV2   & 60.8          & 60.89         & 71.05       & 60.62          &  &  & 85.57         & 88.62         & 95.65       & 84.72          \\
MobileNetV2    & 61.63         & 59.95         & 61.16       & 58.81          &  &  & 85.28         & 84.52         & 87.21       & 83.77          \\
EfficientNetV2 & 56.39         & 56.79         & 79.1        & 44.86          &  &  & 84.58         & 87.6          & 94.85       & 83.04          \\ \hline
\end{tabular}
\end{table*}

\indent The BMFNet-T model guides the BMFNet-S model with the following distillation loss function.
\begin{equation}
{L_{soft}} = Los{s_1} + Los{s_2} + Los{s_3} + Los{s_4} + Los{s_5}
\end{equation}

\section{Results and discussion}

\subsection{Baseline Method}
In this work, the results of state-of-the-art monomodal data mining methods AlexNet [32], RestNet18 [33], EEGNet [35], VGG11 [36], DenseNet [37], ShuffleNetV2 [38], MobileNetV2 [39], and EfficientNetV2 [40] for olfactory EEG and E-nose signals recognition are discussed. In addition, the state-of-the-art multimodal data mining methods M2NN [18], MBERT[28], MulT [29], ViLT [30], and MMASleepNet [31] are compared
with the proposed olfactory EEG and E-nose multimodal learning method.

\subsection{Hyperparameter and Evaluation Index}
This work conducts 24 independent classification experiments for each method, with the samples of each subject as the test set and the other subjects as the training set in turn. The batch size of the training set is 240. The training epoch is 100. Adam optimizer is used for model optimization, with a learning rate of 0.00005 and a weight decay factor of 0.001. The A is set to 0.3 to optimize the , and the distillation temperature T is set to 3. In addition, the accuracy, F1-score, recall, and precision of the model in the olfactory preference recognition task are taken as evaluation metrics to evaluate the model performance comprehensively.

\subsection{Multimodal knowledge distillation of BMFNet}
Table III shows the classification performance and model complexity for the BMFNet-T and BMFNet-S models in the cross-subject olfactory preference recognition task. The BMFNet-S model has more than 40% fewer params and flops than the BMFNet-T model, and their recognition performance is not significantly different. Thus, the proposed knowledge distillation method effectively improves the recognition efficiency of the model while ensuring the classification performance of the model.

\subsection{Monomodal Analysis}

\begin{table}[htbp]
  \centering
  \caption{COMPARISON OF OUR PROPOSED METHOD WITH STATE-OF-THE-ART MULTIMODAL RECOGNITION METHOD}
    \begin{tabular}{p{4.5em}cccc}
    \toprule
    \multirow{2}[2]{*}{Method} & \multicolumn{1}{p{4.19em}}{Accuracy } & \multicolumn{1}{p{4.19em}}{F1-score } & \multicolumn{1}{p{4.19em}}{Recall } & \multicolumn{1}{p{4.19em}}{Precision} \\
    \multicolumn{1}{c}{} & \multicolumn{1}{p{4.19em}}{(\%)} & \multicolumn{1}{p{4.19em}}{(\%)} & \multicolumn{1}{p{4.19em}}{(\%)} & \multicolumn{1}{p{4.19em}}{(\%)} \\
    \midrule
    MMASleepNet & 79.69 & 77.59 & 76.17 & 79.66 \\
    mBERT & 90.16 & 90.79 & 91.82 & 90.19 \\
    M2NN  & 90.21 & 86.67 & 88.21 & 85.63 \\
    MulT  & 90.73 & 90.84 & 91.35 & 90.45 \\
    ViLT  & 90.85 & 90.24 & 90.77 & 89.98 \\
    BMFNet-S & 92.79 & 92.64 & 93.26 & 92.42 \\
    \bottomrule
    \end{tabular}%
  \label{tab:addlabel}%
\end{table}%

\subsubsection{Monomodal Method Comparison}  \indent
\\\indent The recognition results of the monomodal model are shown in Table IV, where RestNet18 and AlexNet are the best for olfactory EEG and E-nose signal recognition, respectively. It can be seen that deeper networks are more advantageous for olfactory EEG recognition. In contrast, the shallower networks may be more suitable for E-nose signal recognition. Furthermore, cross-subject olfactory preference recognition using olfactory EEG is difficult due to individual differences. Its recognition accuracy and other metrics are basically below 70%. For the E-nose samples, the slight differences in feature distributions between samples of the same class give it a better
cross-sample ability. However, E-nose signals have limitations in representing individual olfactory preferences. Even though the overall recognition effect is better than olfactory EEG, the olfactory preferences of some specific subjects are misidentified.

\begin{figure*}[t]
\centering
\includegraphics[width=130mm]{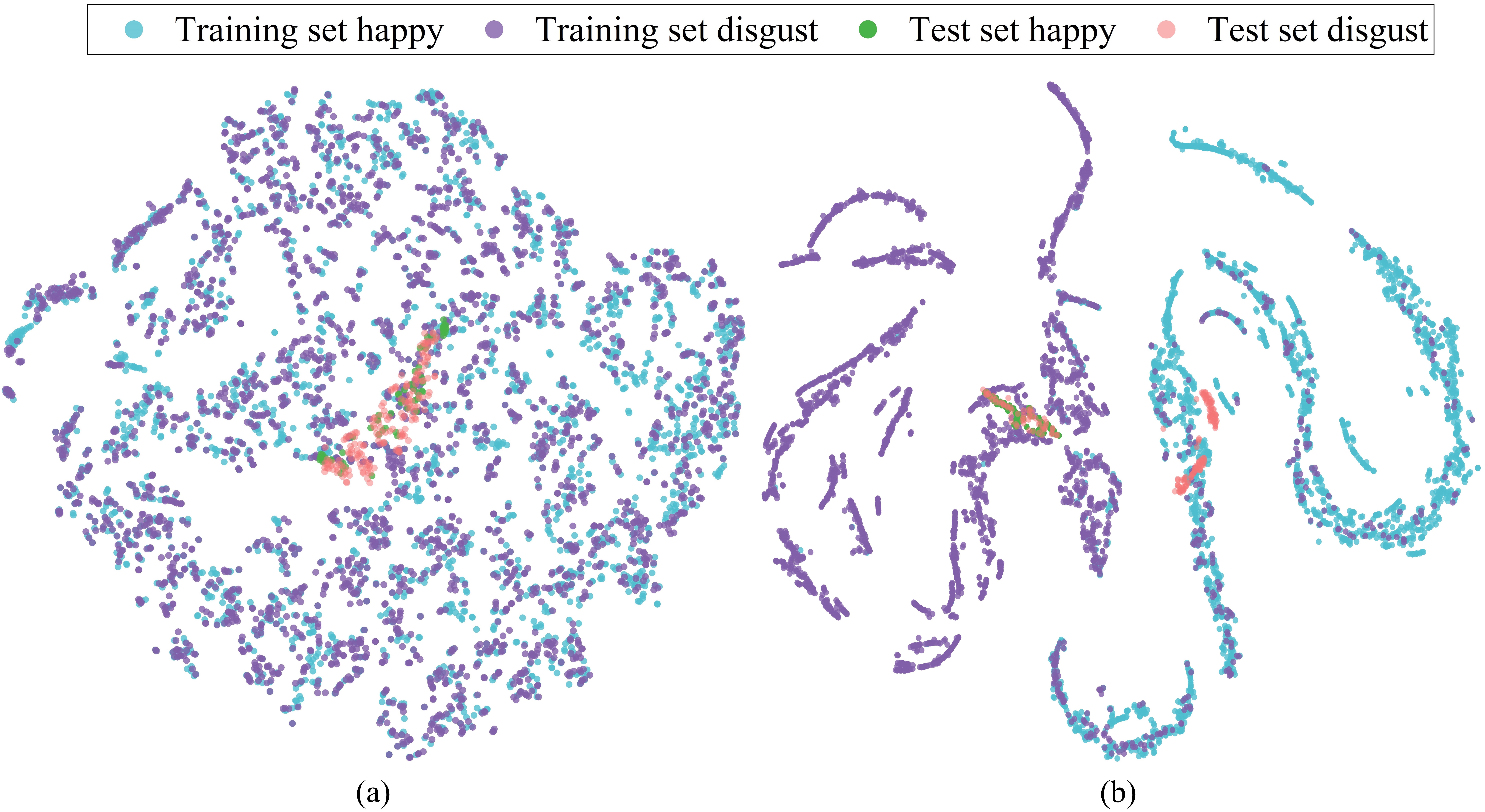}
\caption{Feature mapping of the state-of-the-art monomodal recognition method at the FC layer using t-SNE projections. Light blue, purple, green, and pink colors represent samples from the training set happy, the training set disgust, the test set happy, and the test set disgust, respectively: (a) RestNet18 for olfactory EEG signals recognition, (b) AlexNet for E-nose signals recognition.
\label{overflow}}
\end{figure*}

\subsubsection{Monomodal Feature Visualization}  \indent
\\\indent The t-distributed stochastic neighbor embedding (t-SNE) method can be used to analyze the effectiveness of feature extraction methods [41]. Using the multimodal samples of subject 7 as the test set and the other subjects’ monomodal samples as the training set, Fig. 9 shows the t-SNE visualization of RestNet18 and AlexNet in olfactory EEG and E-nose signals recognition tasks, respectively. It can be seen that the E-nose features are more separable than the olfactory EEG features. When using olfactory EEG for cross-subject olfactory preference recognition, the individual differences and the high complexity of olfactory EEG make it difficult for the model to mine the common and individual features reflecting odor preferences.
\\\indent On the contrary, the spatial difference of E-nose features between samples of the same odor is slight, so it can be said that the E-nose features reflect the common features of the same odor. In addition, as shown in Fig. 9 (b), a few samples in both the training and test sets are misclassified, and their feature spaces are very similar to another class. Because the E-nose can only reflect the odor feature, but not the individual’s preference for the odor. When each subject’s olfactory preference is used as the label for their corresponding E-nose samples, it will likely result in different labels for E-nose signals of the same odor. In this case, the trained model continues to classify based on odor features, which will misclassify some specific olfactory preference samples.

\subsection{Multimodal Analysis}

\subsubsection{Multimodal Method Comparison}  \indent
\\\indent The proposed method is compared with state-of-the-art multimodal data mining methods to evaluate the proposed method’s performance. As shown in Table V, the accuracy, F1-score, recall, and precision of the proposed method are the most competitive compared to the state-of-the-art methods.

\begin{figure*}[t]
\centering
\includegraphics[width=183mm]{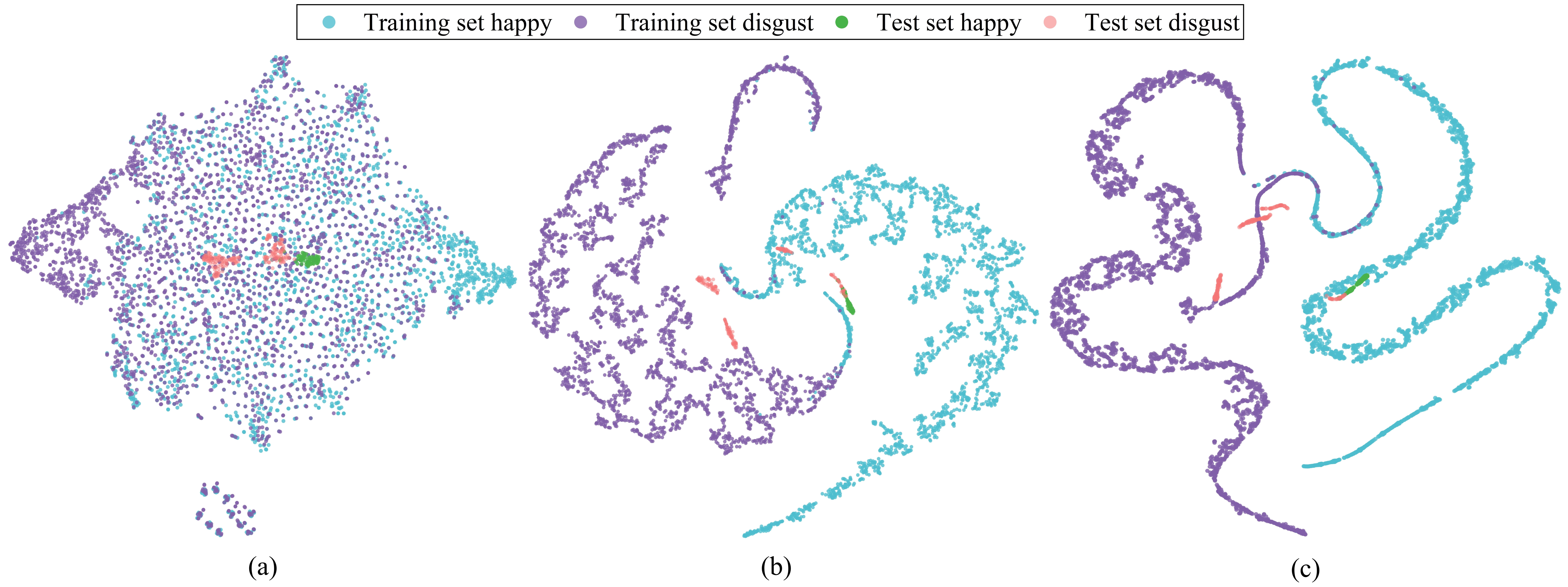}
\caption{Feature mapping of the state-of-the-art multimodal recognition method at the FC layer using t-SNE projections. Light blue, purple, green, and pink colors represent samples from the training set happy, the training set disgust, the test set happy, and the test set disgust, respectively: (a) ViLT, (b) MulT, and (c) BMFNet-S.
\label{overflow}}
\end{figure*}

\subsubsection{Multimodal Feature Visualization}  \indent
\\\indent Using the multimodal samples of subject 7 as the test set and the other subjects’ multimodal samples as the training set, Fig. 10 (a), Fig. 10 (b), and Fig. 10 (c) show the t-SNE visualization of the ViLT, MulT, and BMFNet-S model in the cross-subject olfactory preference recognition task, respectively. It can be seen that the ViLT model establishes the feature space of multimodal samples by inputting the embedded olfactory EEG and E-nose data into the self-attention encoder in parallel. Still, it is difficult to mine the features representing individual olfactory preferences from olfactory EEG data. For the MulT model, the cross-modal attention mechanism realizes the cross-modal transfer between different modal information, and the interaction between the multimodal sequences well mines the individual olfactory preference features. However, the established feature space is not separable in the training set, and a few samples are still misclassified in the test set. The proposed BMFNet-S model establishes a suitable feature space of multimodal samples, effectively separating the different classes on the training and test sets. Compared with other state-of-the-art multimodal data mining methods, the proposed method has significant advantages in mining individual olfactory preference features.

\begin{figure*}[t]
\centering
\includegraphics[width=183mm]{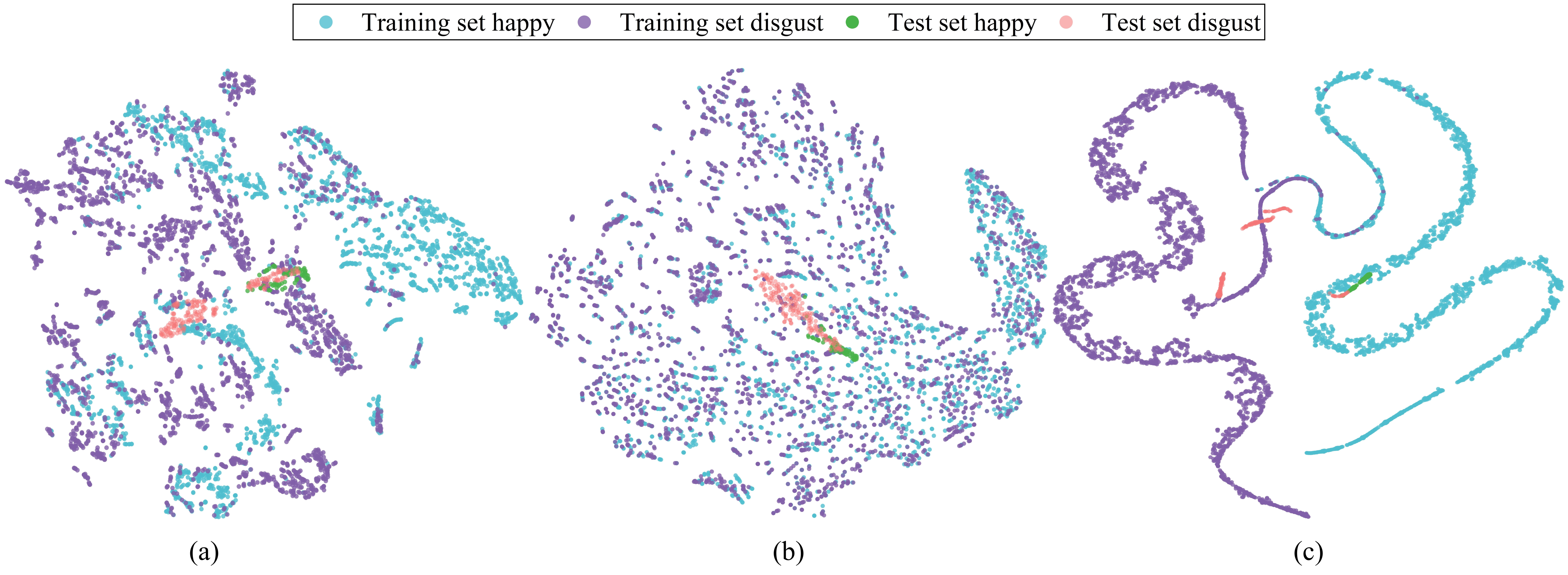}
\caption{Feature mapping of each module within the proposed method using t-SNE projections. Light blue, purple, green, and pink colors represent samples from the training set happy, the training set disgust, the test set happy, and the test set disgust, respectively: (a) AEFM, (b) MFI, and (c) FF.
\label{overflow}}
\end{figure*}

\begin{figure*}[t]
\centering
\includegraphics[width=183mm]{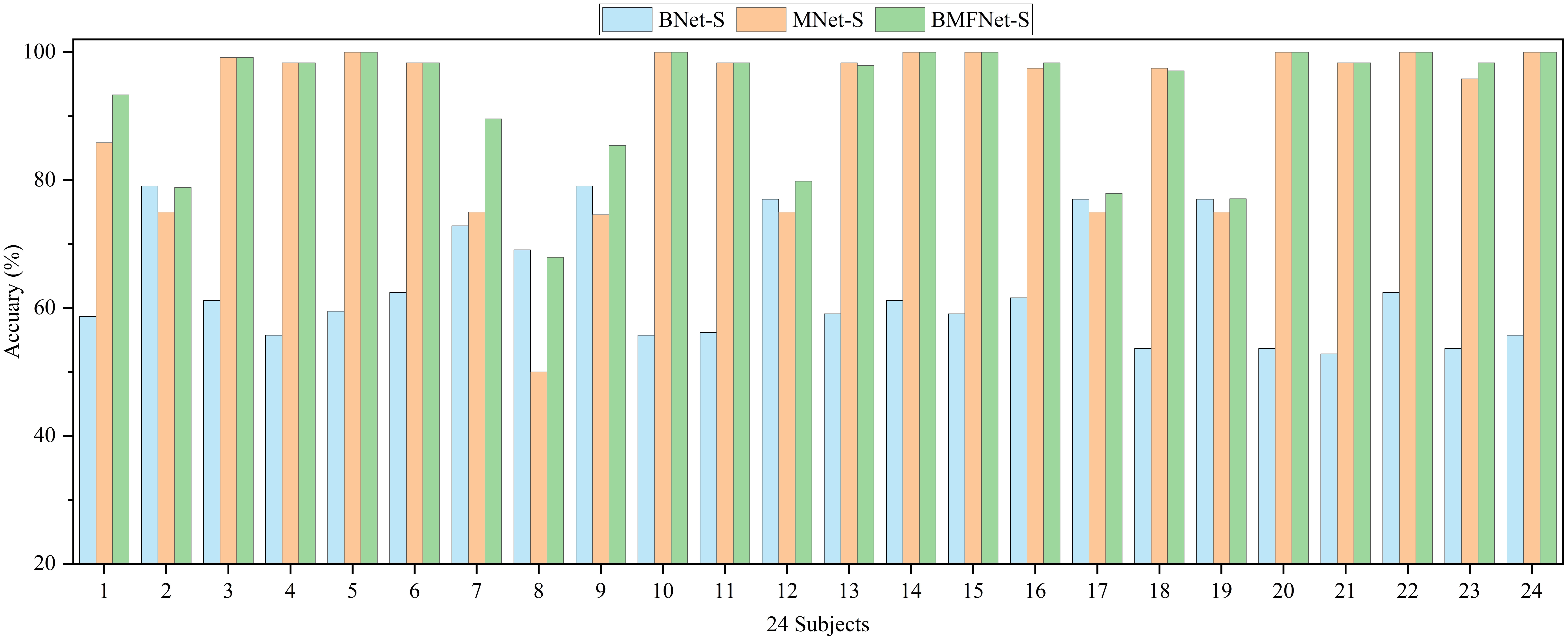}
\caption{The classification results of each Bnet-S, MNet-S, and BMFNet-S parallel experiment.
\label{overflow}}
\end{figure*}

\subsection{Ablation Experiment of BMFNet-S}
Detailed ablation studies are performed to verify the proposed method’s architecture. Firstly, the validity of the AEFM, MFI, and FF modules is initially verified through the visualization of their output features. Secondly, the MFI and AlexNet modules in BMFNet-S are removed to build a BNet-S model. The MFI and RestNet modules in BMFNet-S are removed to build an MNet-S model. By comparing the results of BNet-S, MNet-S, and BMFNet-S, it is further verified the significance of mining the common features between olfactory EEG and E-nose
signals through MFI. Finally, the effectiveness of contrastive loss and AEFM modules is further discussed.

\begin{figure*}[t]
\centering
\includegraphics[width=183mm]{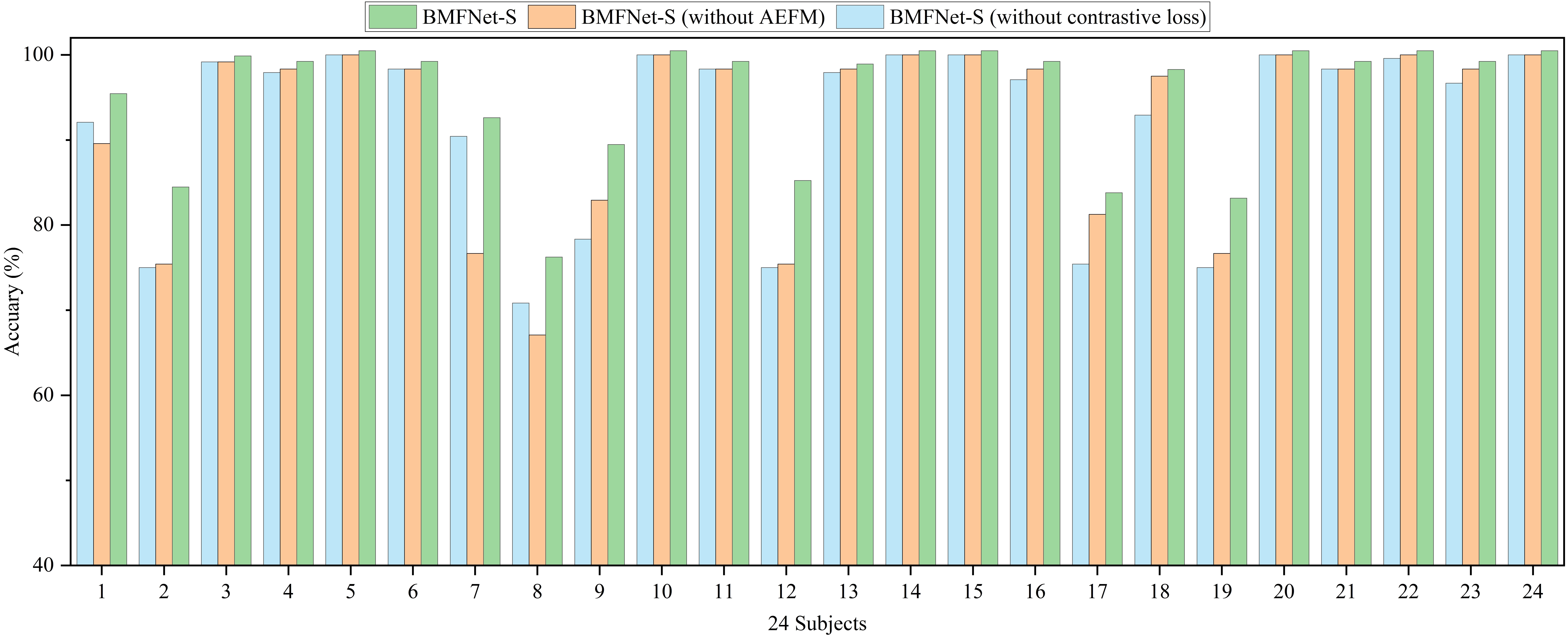}
\caption{The classification results of BMFNet-S, BMFNet-S (without AEFM), and BMFNet-S (without contrastive loss) models.
\label{overflow}}
\end{figure*}

\indent The output features visualization of AEFM, MFI, and FF modules in BMFNet-S are shown in Fig. 11 (a), Fig. 11 (b), and Fig. 11 (c), respectively. The feature space of the test set samples in Fig. 11 (a) shows that the AEFM module mines the individual features that represent the olfactory preference of subject 7. Nevertheless, many training set samples of different classes in the feature space are difficult to separate, suggesting its limitations in overcoming the problem of individual differences. The feature space of the samples in Fig. 11 (b) shows that the MFI module effectually exploits the common features between the olfactory EEG and E-nose signals. It solves the problem of difficulty in cross-subject recognition due to individual differences to some extent. Still, by observing the feature space of the test set samples, it can be found that some samples still have poorly distinguishable features. The feature space of the samples in Fig. 11 (c) shows that the FF module effectively fuses the common features of the olfactory EEG and E-nose signals as well as the individual features in the olfactory EEG, which enables the BMFNet-S model to realize cross-subject olfactory preference recognition effectively.

\indent Fig. 12 shows the classification results of BNet-S, MNet-S, and BMFNet-S models. It can be seen that the recognition effect of the BNet-S model is worse than that of MNet-S because cross-subject recognition is difficult. The MNet-S model only completes the training according to the preferences of most subjects in the training set, and it essentially does not have the cross-subject olfactory preference recognition ability. So, serious recognition errors will occur when recognizing subjects with special odor preferences, such as subjects 7, 8, and 9. The proposed BMFNet-S model effectively overcomes the difficulty of individual olfactory EEG differences by mining the common features between olfactory EEG and E-nose signals. Finally, the model fuses common features that ensure cross-subject recognition ability and individual features that ensure olfactory preference recognition ability to comprehensively and accurately represent each person’s olfactory preference.

Figure 13 shows the classification results of BMFNet-S, BMFNet-S (without AEFM), and BMFNet-S (without contrastive loss) models. By comparing BMFNet-S with BMFNet-S (without AEFM), it can be seen that the introduction of the AEFM module effectively improves the model’s cross-subject olfactory preference recognition ability. By comparing BMFNet-S and BMFNet-S (without contrastive loss), it can be seen that the introduction of contrast loss effectively improves the feature distribution of olfactory EEG samples and avoids the recognition difficulties caused by individual olfactory EEG differences as much as possible.

\subsection{Limitations and Potential Applications}
In this study, we tried our best to establish a data set with 5760 multimodal samples (24 subjects and 4 odors), and the effectiveness of the proposed method has been preliminarily proved. Still, we need to introduce more subjects and odor types to make the research closer to the practical application of odor sensory evaluation. In addition, the number of sensors in the E-nose used is insufficient compared to the human olfactory receptors, thus making it difficult to establish the interaction between bionic and human olfaction fully. In the future, we will improve the number and types of sensors in the E-nose to make our method better serve odor sensory evaluation tasks.
\\\indent Such sleep, auditory, and motor imagery EEG tasks also face the problem of cross-subject EEG recognition, and our research provides ideas for solving this problem. In these EEG tasks, modalities similar to the E-nose, which reflect the objective properties of the task, can be combined with the subject's EEG to achieve cross-subject EEG recognition. Therefore, this method has a good application prospect in cross-subject EEG recognition.

 \vspace{-0.08cm}%%压缩图片和cap之间
 
\section{Conclusion}
In this paper, an olfactory EEG and E-nose multimodal learning method is proposed for cross-subject olfactory preference recognition. It can sufficiently establish the interaction between bionic and human olfaction to mine odor information and human emotions. A complementary multimodal data mining strategy is established to effectively mine the common features between olfactory EEG and E-nose and the individual features in olfactory EEG. Then, common and individual features are fused and used for classification. The experimental results show the proposed method can effectively recognize cross-subject olfactory preference, and it outperforms state-of-the-art recognition methods, which shows it is potentially valuable for odor sensory evaluation.


\begin{thebibliography}{00}
\bibitem{ref1} S. Lombion-Pouthier, P. Vandel P, S. Nezelof, E. Haffen and J. L. Millot, “Odor perception in patients with mood disorders,” Journal of Affective Disorders, vol. 90, no. 2-3, pp. 187-191, Feb. 2006, doi:10.1016/j.jad.2005.11.012.
\bibitem{ref2} A. Wrzesniewski, C. McCauley, P. Rozin, “Odor and affect: individual differences in the impact of odor on liking for places, things and people,” Chemical Senses, vol. 24, no. 6, pp. 713-721, Dec. 1999, doi:10.1093/chemse/24.6.713.
\bibitem{ref3} E, Mc Donnell, S. Hulin-Bertaud, E. M. Sheehan and C. M. Delahunty, “Development and learning process of a sensory vocabulary for the odor evaluation of selected distilled beverages using descriptive analysis,” Journal of Sensory Studies, vol. 16, no. 4, pp. 425-445, Aug. 2001, doi:10.1111/j.1745-459X.2001.tb00311.x.
\bibitem{ref4} R. H. McQueen and S. Vaezafshar, “Odor in textiles: A review of evaluation methods, fabric characteristics, and odor control technologies,” Textile Research Journal, vol. 90, no. 9-10, pp. 1157-1173, May. 2020, doi:10.1177/0040517519883952. 
\bibitem{ref5} M.A. Jeltema and E. W. Southwick, “Evaluation and applications of odor profiling,” Journal of Sensory Studies, vol. 1, no. 2, pp. 123-136, Jun. 1986, doi:10.1111/j.1745-459X.1986.tb00165.x.

\bibitem{ref6} M. Verriele, H. Plaisance, V. Vandenbilcke, N. Locoge, J. N. Jaubert and G. Meunier, “Odor evaluation and discrimination of car cabin and its components: Application of the ‘field of odors’ approach in a sensory descriptive analysis,” Journal of Sensory Studies, vol. 27, no. 2, pp. 102-110, Apr. 2012, doi:10.1111/j.1745-459X.2012.00371.x.
\bibitem{ref7} J. Y. Zhang et al., “A miniaturized electronic nose with artificial neural network for anti-interference detection of mixed indoor hazardous gases,” Sensors and Actuators B-Chemical, vol. 326, Jan. 2021, Art. no. 128822, doi:10.1016/j.snb.2020.128822.
\bibitem{ref8} Z. Haddi, A. Amari, H. Alami, N. El Bari, E. Llobet and B. Bouchikhi, “A portable electronic nose system for the identification of cannabis-based drugs,” Sensors and Actuators B-Chemical, vol. 155, no. 2, pp. 456-463, Jul. 2011, doi:10.1016/j.snb.2010.12.047.
\bibitem{ref9} B. Liu et al., “Lung cancer detection via breath by electronic nose enhanced with a sparse group feature selection approach,” Sensors and Actuators B-Chemical, vol. 339, Jul. 2021, Art. no. 129896, doi:10.1016/j.snb.2021.129896.
\bibitem{ref10} D. Liu, W. Dai, H. Zhang, X. Jin, J. Cao and W. Kong, “Brain-machine coupled learning method for facial emotion recognition,” IEEE Trans. Pattern Anal. Mach, Intell., vol. 45, no. 9, pp. 10703-10717, Sep. 2023, doi:10.1109/TPAMI.2023.3257846.
\bibitem{ref11} S. Issa, Q. Peng, and X. You, “Emotion Classification Using EEG Brain Signals and the Broad Learning System,” IEEE Trans. Syst. Man Cybern, Syst., vol. 51, no. 12, pp. 7382–7391, Dec. 2021, doi: 10.1109/TSMC.2020.2969686.
\bibitem{ref12} W.-L. Zheng, W. Liu, Y. Lu, B.-L. Lu, and A. Cichocki, “EmotionMeter: A Multimodal Framework for Recognizing Human Emotions,” IEEE Trans. Cybern., vol. 49, no. 3, pp. 1110–1122, Mar. 2019, doi: 10.1109/TCYB.2018.2797176.


\bibitem{ref13} X. X. Xia, X. T. Liu, W. B. Zheng, X. F. Jia, B. Wang, Y. Shi and H. Men, “Recognition of odor and pleasantness based on olfactory EEG combined with functional brain network model,” International Journal Of Machine Learning and Cybernetics, vol. 14, no. 8, pp. 2761-2776, Aug. 2023, doi:10.1007/s13042-023-01797-7.
\bibitem{ref14} Z. Gao, W. Dang, M. Liu, W. Guo, K. Ma, and G. R. Chen, “Classification of EEG Signals on VEP-Based BCI Systems With Broad Learning,” IEEE Trans. Syst. Man Cybern, Syst., vol. 51, no. 11, pp. 7143–7151, Nov. 2021, doi: 10.1109/TSMC.2020.2964684.
\bibitem{ref15} X. Xia, M. Wang, Y. Shi, Z. Huang, J. Liu, H. Men and H. Fang, “Identification of white degradable and non-degradable plastics in food field: A dynamic residual network coupled with hyperspectral technology,” Spectrochimica Acta Part A Molecular and Biomolecular Spectroscopy, vol. 296, pp. 122686, Aug. 2023, doi:10.1016/j.saa.2023.122686.
\bibitem{ref16} H. L. Lin, H. M. Chen, C. B. Yin, Q. L. Zhang, Z. Y. Li, Y. Shi and H. Men, “Lightweight residual convolutional neural network for soybean classification combined with electronic nose,” IEEE Sensors Journal, vol. 22, no. 12, pp. 11463-11473, Jun. 2022, doi:10.1109/JSEN.2022.3174251.
\bibitem{ref17} X. X. Xia, Y. Shi, P. Li, X. Liu, J. Liu and H. Men, “FBANet: An Effective Data Mining Method for Food Olfactory EEG Recognition,” IEEE Trans. Neural Netw. Learn Syst., May. 2023, early access, doi:10.1109/TNNLS.2023.3269949.

\bibitem{ref18} Q. He, L. F. Feng, G. Q. Jiang and P. Xie, “Multimodal multitask neural network for motor imagery classification with EEG and fNIRS signals,” IEEE Sensors Journal, vol. 22, no. 21, pp. 20695-20706, Nov. 2022, doi:10.1109/JSEN.2022.3205956.
\bibitem{ref19} S. Buratti, C. Malegori, S. Benedetti, P. Oliveri and G. Giovanelli, “E-nose, e-tongue and e-eye for edible olive oil characterization and shelf life assessment: A powerful data fusion approach,” Talanta, vol. 182, pp. 131-141, May. 2018, doi:10.1016/j.talanta.2018.01.096.
\bibitem{ref20} H. L. Ma et al., “A low-cost and efficient electronic nose system for quantification of multiple indoor air contaminants utilizing HC and PLSR,” Sensors and Actuators B-Chemical, vol. 350, Jan. 2022, Art. no. 130768, doi:10.1016/j.snb.2021.130768.
\bibitem{ref21} X. W. Peng, L. Zhang, F. C. Tian and D. Zhang, “A novel sensor feature extraction based on kernel entropy component analysis for discrimination of indoor air contaminants,” Sensors and Actuators A Physical, vol. 234, pp. 143-149, Oct. 2015, doi:10.1016/j.sna.2015.09.009.
\bibitem{ref22} L. Zhang and D. Zhang, “Domain adaptation extreme learning machines for drift compensation in E-nose systems,” IEEE Transactions on Instrumentation and Measurement, vol. 64, no. 7, pp. 1790-1801, Jul. 2015, doi:10.1109/TIM.2014.2367775.

\bibitem{ref23} P. Peng, X. J. Zhao, X. F. Pan and W. B. Ye, “Gas classification using deep convolutional neural networks,” Sensors, vol. 18, no. 2, Jan. 2018, Art. no. 157, doi:10.3390/s18010157.
\bibitem{ref24} L. H. Feng, H. H. Dai, X. Song, J. M. Liu and X. Mei, “Gas identification with drift counteraction for electronic noses using augmented convolutional neural network,” Sensors and Actuators B-Chemical, vol. 351, Jan. 2022, Art. no. 130986, doi:10.1016/j.snb.2021.130986.
\bibitem{ref25} X. J. Zhao, Z. H. Wen, X. F. Pan, W. B. Ye and A. Bermak, “Mixture gases classification based on multi-label one-dimensional deep convolutional neural network,” IEEE Access, vol. 7, pp. 12630-12637, Feb. 2019, doi:10.1109/ACCESS.2019.2892754.
\bibitem{ref26} K. Ezzatdoost, H. Hojjati and H. Aghajan, “Decoding olfactory stimuli in EEG data using nonlinear features: A pilot study,” Journal of Neuroscience Methods, vol. 341, Jul. 2020, Art. no. 108780, doi:10.1016/j.jneumeth.2020.108780.
\bibitem{ref27} O. Aydemir, “Odor and subject identification using electroencephalography reaction to olfactory,” Traitement Du Signal, vol. 37, no. 5, pp. 799-805, Nov. 2020, doi:10.18280/ts.370512.

\bibitem{ref28} J. F. Yu and J. Jiang, “Adapting BERT for target-oriented multimodal sentiment classification,” in Proceedings of the Twenty-Eighth International Joint Conference on Artificial Intelligence, Macao, PEOPLES R CHINA, 2019, pp. 5408-5414, doi:10.24963/ijcai.2019/751.
\bibitem{ref29} Y. H. H. Tsai, S. J. Bai, P. P. Liang, J. Z. Kolter, L. P. Morency and R. Salakhutdinov, “Multimodal transformer for unaligned multimodal language sequences,” in 57th Annual Meeting of the Association for Computational Linguistics (ACL 2019), Stroudsburg, PA, USA, 2019, pp. 6558-6569, doi:10.18653/v1/p19-1656.
\bibitem{ref30} W. Kim, B. Son and I. Kim, “ViLT: Vision-and-language transformer without convolution or region supervision,” in International Conference on Machine Learning, ELECTR NETWORK, vol. 139, 2021.
\bibitem{ref31} Z. Yubo, L. Yingying, Z. Bing, Z. Lin and L. Lei, “MMASleepNet: A multimodal attention network based on electrophysiological signals for automatic sleep staging,” Frontiers in Neuroscience, vol. 16, Aug. 2022, Art. no. 973761, doi:10.3389/fnins.2022.973761.
\bibitem{ref32} A. Krizhevsky, I. Sutskever, and G. E. Hinton, “Imagenet classification with deep convolutional neural networks,” Advances in neural information processing systems, vol. 25, pp. 1097–1105, 2012. 3, 9, doi:10.1145/3065386. 

\bibitem{ref33} K. He, X. Zhang, S. Ren, and J. Sun, “Deep residual learning for image recognition,” in Proceedings of the IEEE conference on computer vision and pattern recognition, pp. 770–778, 2016. 3, 9, doi:10.1109/CVPR2016.90.
\bibitem{ref34} D. A. Schmidt, C. Shi, R. A. Berry, M. L. Honig and W. Utschick, “Minimum mean squared error interference alignment,” in 2009 Conference Record of the Forty-Third Asilomar Conference on Signals Systems and Computers, Pacific Grove, CA, USA, 2009, pp. 1106-1110, doi:10.1109/ASILOMAR15230.2009.
\bibitem{ref35} V. J. Lawhern, A. J. Solon, N. R. Waytowich, S. M. Gordon, C. P. Hung and B. J. Lance, “EEGNet: a compact convolutional neural network for EEG-based brain-computer interfaces,” J. Neural Eng., vol. 15, no. 5, Oct. 2018, Art. no. 056013, doi:10.1088/1741-2552/aace8c.
\bibitem{ref36} K. Simonyan and A. Zisserman, “Very deep convolutional networks for large-scale image recognition,” Computer Science, 2014, doi:10.48550/arXiv.1409.1556.
\bibitem{ref37} G. Huang, Z. Liu, L. van der Maaten and K. Q. Weinberger, “Densely connected convolutional networks,” in Proceedings of the IEEE Conference on Computer Vision and Pattern Recognition (CVPR), 2017, pp. 4700-4708, doi:10.48550/arXiv.1608.06993.

\bibitem{ref38} N. Ma, X. Zhang, H.-T. Zheng, and J. Sun, “Shufflenet v2: Practical guidelines for efficient cnn architecture design,” in Proceedings of the European conference on computer vision (ECCV),pp. 116–131, 2018, doi:10.1007/978-3-030-01264-9\_8.
\bibitem{ref39} M. Sandler, A. Howard, M. L. Zhu, A. Zhmoginov and L. C. Chen, “MobileNetV2: Inverted residuals and linear bottlenecks,” in 2018 IEEE/CVF Conference on Computer Vision and Pattern Recognition (CVPR), Salt Lake City, UT, 2018, pp. 4510-4520, doi: 10.1109/CVPR.2018.00474.
\bibitem{ref40} M. X. Tan and Q. V. Le, “EfficientNetV2: Smaller models and faster training,” in International Conference on Machine Learning, ELECTR NETWORK, vol. 139, 2021, pp. 7102-7110.
\bibitem{ref41} L. van der Maaten, and G. Hinton, “Visualizing data using t-SNE,” Journal of Machine Learning Research, vol. 9, pp. 2579-2605, Nov. 2008, doi:10.48550/arXiv.2108.01301.


\end{thebibliography}
\end{document}